\RequirePackage{fix-cm}
\documentclass[twocolumn]{svjour3}          
\smartqed  
\usepackage{graphicx}
\usepackage{natbib}
\usepackage{multirow}
\usepackage{amsmath}
\usepackage[none]{hyphenat}
\usepackage[misc]{ifsym}

\usepackage{color}

\begin{document}

\title{3D-FUTURE: 3D Furniture shape with TextURE 
}


\author{Huan Fu         \and
        Rongfei Jia \and 
        Lin Gao \and
        Mingming Gong \and
        Binqiang Zhao \and
        Steve Maybank \and
        Dacheng Tao
}


\institute{
Huan Fu (\Letter) \at 
TaoXi Technology Department, Alibaba Group, CN \\
\email{fuhuan.fh@alibaba-inc.com} \and
Rongfei Jia \at 
TaoXi Technology Department, Alibaba Group, CN \\
\email{rongfei.jrf@alibaba-inc.com} \and
Lin Gao \at 
Institute of Computing Technology, Chinese Academy of Sciences, CN \\
\email{gaolinorange@gmail.com} \and
Mingming Gong \at 
The University of Melbourne, VIC, AU \\
\email{mingming.gong@unimelb.edu.au} \and
Binqiang Zhao \at 
TaoXi Technology Department, Alibaba Group, CN \\
\email{binqiang.zhao@alibaba-inc.com} \and
Steve Maybank \at 
Department of Computer Science and Information Systems, Birkbeck College, University of London, UK \\
\email{sjmaybank@dcs.bbk.ac.uk} \and
Dacheng Tao \at
UBTECH Sydney AI Centre, The University of Sydney, NSW, AU \\
\email{dacheng.tao@sydney.edu.au}
}

\date{Received: date / Accepted: date}

\maketitle

\begin{abstract}
\sloppy
The 3D CAD shapes in current 3D benchmarks are mostly collected from online model repositories. Thus, they typically have insufficient geometric details and less informative textures, making them less attractive for comprehensive and subtle research in areas such as high-quality 3D mesh and texture recovery. This paper presents 3D Furniture shape with TextURE (3D-FUTURE): a richly-annotated and large-scale repository of 3D furniture shapes in the household scenario. At the time of this technical report, 3D-FUTURE contains 20,240 clean and realistic synthetic images of 5,000 different rooms. There are 9,992 unique detailed 3D instances of furniture with high-resolution textures. Experienced designers developed the room scenes, and the 3D CAD shapes in the scene are used for industrial production. Given the well-organized 3D-FUTURE, we provide baseline experiments on several widely studied tasks, such as joint 2D instance segmentation and 3D object pose estimation, image-based 3D shape retrieval, 3D object reconstruction from a single image, and texture recovery for 3D shapes, to facilitate related future researches on our database.

\end{abstract}

\section{Introduction}
\label{sec:intro}
\sloppy
The rapid progress of modern machine learning methods, such as deep neural models, has led to various impressive breakthroughs towards 2D computer vision and natural language processing (NLP). One key to facilitating the advancement of these approaches is the availability of large-scale labeled benchmarks. 
Mirroring this pattern, the computer graphics and 3D vision communities have put tremendous efforts in establishing 3D datasets over the past years, expecting to enable and innovate the avenues of future research \citep{chang2015shapenet,wu20153d,xiao2013sun3d,song2015sun,xiao2016sun,sun2018pix3d,xiang2014beyond,xiang2016objectnet3d,silberman2012indoor,dai2017scannet,hua2016scenenn}. For example, the largest 3D repositories, like ShapeNet \citep{chang2015shapenet} and ModelNet \citep{wu20153d}, collected massive 3D shapes from online repositories and organized them under the WordNet taxonomy. Relying on the repositories, several works, such as Pascal 3D+ \citep{xiang2014beyond}, ObjectNet3D \citep{xiang2016objectnet3d}, Pix3D \citep{sun2018pix3d}, and Stanford Cars \citep{KrauseStarkDengFei-Fei_3DRR2013}, further provided images and shapes associations or alignments with fine-grained pose annotations. Other works like NYU Depth Dataset \citep{silberman2012indoor}, SUN RGB-D \citep{song2015sun}, ScanNet \citep{dai2017scannet}, SceneNN \citep{hua2016scenenn}, and Matterport3D \citep{chang2017matterport3d} introduced RGB-D scans of real-world indoor environments with many estimated and manually verified annotations. Considering that there are rich 3D benchmarks, why do we need one more?

\setlength\tabcolsep{2.5pt}
\begin{table*}[h]
\centering
\small
\begin{tabular}{ c || c  c  c  c | c  c  c }
\hline
Benchmarks & Shapes & \textbf{Texture} & Categories & \textbf{Shape Source} & Scene Images & Instances & \textbf{Alignments} \\
\hline
PrincetonSB \citep{shilane2004princeton} & 6,670 & $\times$ & 161 & Online & $\times$ & $\times$ & $\times$ \\
ShapeNetCore \citep{chang2015shapenet} & 51,300 & $\surd^{*}$ & 55 & Online & $\times$ & $\times$ & $\times$ \\
ShapeNetSem \citep{chang2015shapenet} & 12,000 & $\surd^{*}$ & 270 & Online & $\times$ & $\times$ & $\times$ \\
ModelNet \citep{wu20153d} & 151,128 & $\times$ & 660 & Online & $\times$ & $\times$ & $\times$ \\
ObjectScans \citep{choi2016large} & $\sim$1,900 & $\times$ & 44 & Scans & $\times$ & $\times$ & $\times$ \\
\hline
IKEA \citep{lim2013parsing} & 219 & $\times$ & 11 & Industry & 759 & - & pseudo \\
PASCAL3D+ \citep{xiang2014beyond} & 79 & $\times$ & 12 & ShapeNet & $\times$ & 30,899 & raw \\
ObjectNet3D+ \citep{xiang2016objectnet3d} & 44,161 & $\times$ & 100 & ShapeNet & 90,127 & 201,888 & raw \\
Pix3D \citep{sun2018pix3d} & 395 & $\surd^{*}$ & 5 & ShapeNet & $\times$ & 10,069 & pseudo \\
Standford Cars \citep{KrauseStarkDengFei-Fei_3DRR2013} & 134 & $\surd^{*}$ & 1 & ShapeNet & $\times$ & 16,185 & pseudo \\
Comp Cars \citep{yang2015large} & 98 & $\surd^{*}$ & 1 & ShapeNet & $\times$ & 5,696 & pseudo \\
ScanNet \citep{dai2017scannet} & 296 & $\surd^{*}$ & 17 & ShapeNet & 1513 scans & $\sim$9,600 & pseudo \\
\hline
InteriorNet \citep{li2018interiornet} & N/A & $\times$ & N/A & N/A & 20M$^\dag$ & $\times$  & $\times$ \\
Structured3D \citep{zheng2019structured3d} & N/A & $\times$ & N/A & N/A & 20M$^\dag$ & $\times$  & $\times$ \\
\hline
3D-FUTURE (ours) & 9,992 & \textcolor{red}{$\surd$} & 34 & \textcolor{red}{Industry} & 20,240$^\dag$ & 102,972  & \textcolor{red}{precise} \\
 \hline
\end{tabular}
\caption{Statistics of some representative 3D benchmarks. Instances: images with saliency objects (like images in ImageNet \citep{krizhevsky2012imagenet}). Alignments: 2D to 3D alignment annotations. $\surd^*$: The shapes are with uninformative textures, and only part of shapes comes with textures. $\sim$: about. $\dag$: synthetic images. ``Raw" and ``pseudo" mean that the 3D shapes are usually not the exact the ones corresponding to the 2D objects. Note that, our 3D-FUTURE is specific to household scenario, and all the 3D shapes are industrial used furniture shapes. See Figure~\ref{fig:real-2d-3d-alignments}, Figure~\ref{fig:shape-texture}, and Figure~\ref{fig:photorealistic-synthetic-images} for more details of our highlight features.}
\end{table*}

In contrast to the 2D counterparts \citep{krizhevsky2012imagenet,lin2014microsoft,geiger2012we}, we realize that there is still a big gap between 3D academic research and industrial productions.  For instance, the 3D CAD models in existing datasets mainly come from public online repositories like Trimble 3D Warehouse\footnote{https://3dwarehouse.sketchup.com} and Yobi3D\footnote{https://yobi3d.com}. These 3D shapes typically have fewer geometry details and uninformative textures or even no textures. Specific to shapes in the household scenario, most of them are outdated and dull furniture deprecated by modern professional designers. Therefore, the current 3D shapes are inadequate for comprehensive and subtle research in areas such as industry closely related fine-grained 3D shape understanding and texture recovery. Besides, existing benchmarks only provide pseudo image or shape alignments, and the estimated camera pose annotations. Namely, the benchmark designers manually choose a roughly matched 3D CAD model from available 3D shape benchmarks according to the object in the image. Thus, annotators may largely ignore some local shape details, which prevents the progress of fundamental data-driven studies such as high-quality 3D reconstruction from real-world images and high-accuracy image-based 3D shape retrieval. Last but not least, there is no well-organized benchmark that offers realistic synthetic indoor images with both instance-level semantic annotations and the involved 3D shapes with textures. 

\begin{figure*}[th]
\centering
\includegraphics[width=0.98\textwidth]{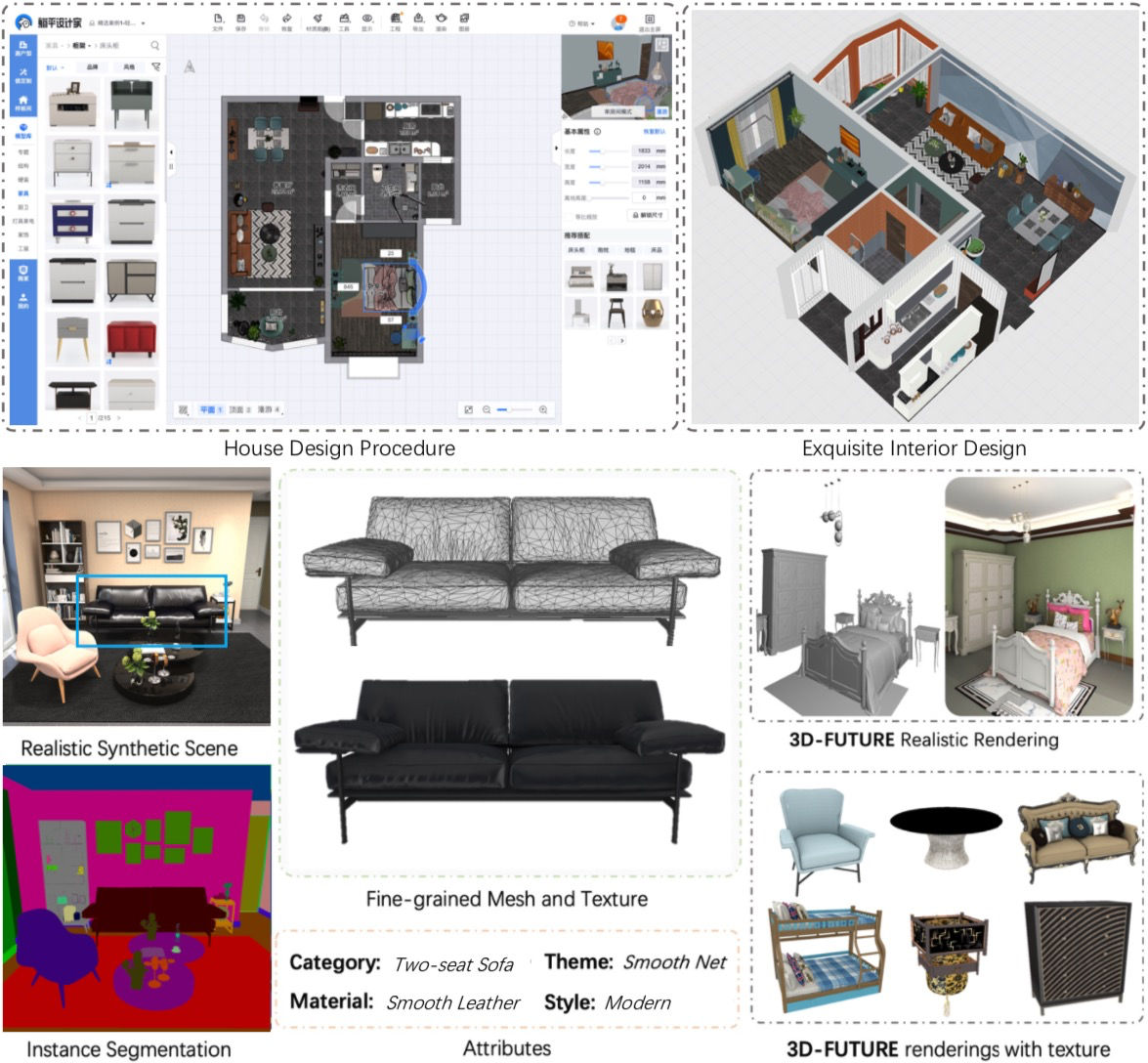}
\caption{\textbf{3D-FUTURE.} Top: Exquisite interior designs obtained from Alibaba Topping Homestyler design platform. Bottom: An overview of the properties of 3D-FUTURE. All the interior designs are developed or reviewed by experienced designers to ensure their quality. The photo-realistic synthetic scenes are rendered by the advanced rendering engine V-ray. The statistics of 3D-FUTURE are presented in Sec.~\ref{sec:properties_statistics}.}
\label{fig:3d-future-overview}
\end{figure*}

\begin{figure*}[th]
\centering
\includegraphics[width=0.95\textwidth]{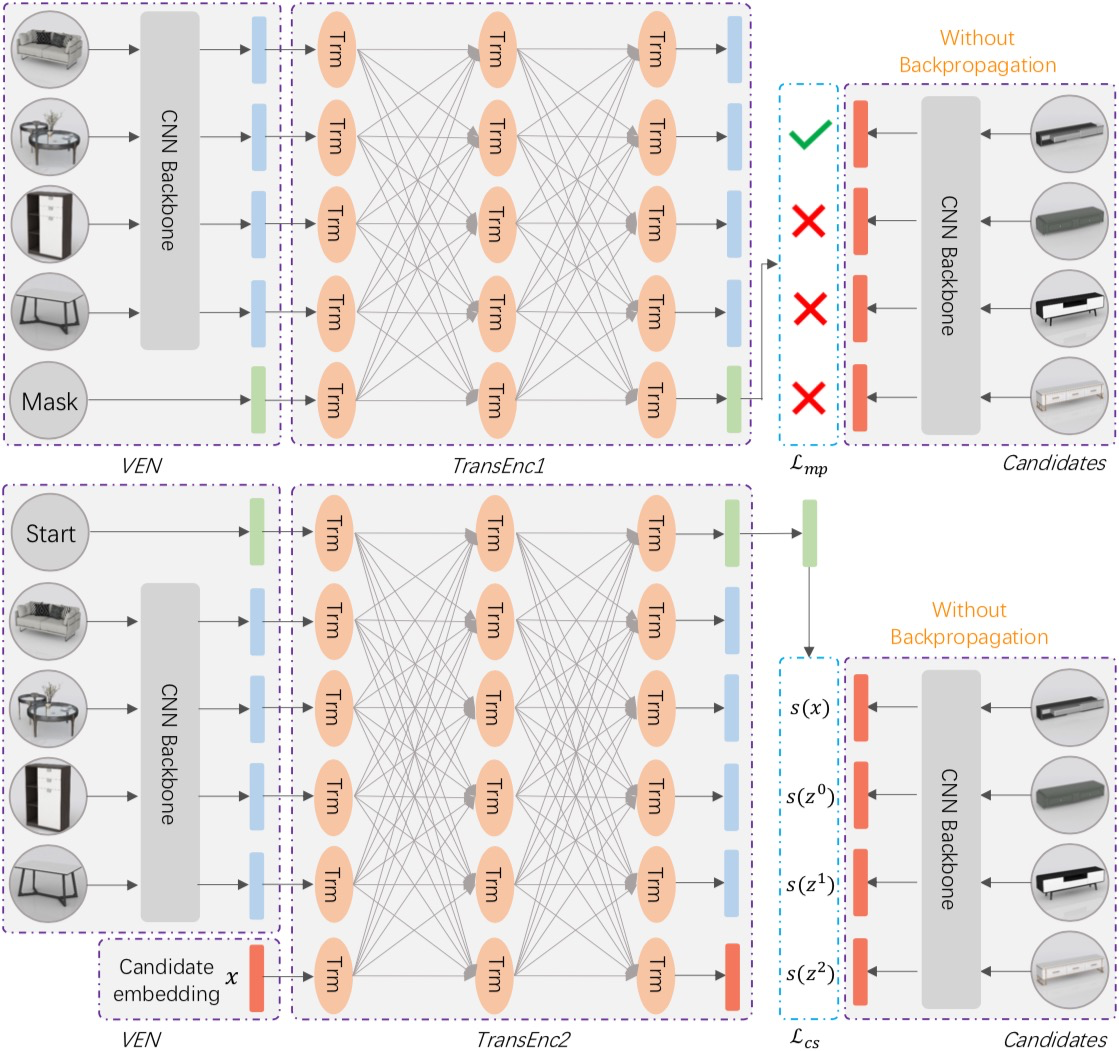}
\caption{\textbf{DFSM.} An illustration of the deep furnishing suit model (DFSM) for deep visual embedding in Sec.~\ref{subsubsec:deep_visual_embedding}. The development of the framework borrows the concepts from Bert \citep{devlin2018bert}. We construct two tasks here, including mask prediction and compatibility scoring, as explained in Sec.~\ref{subsubsec:deep_visual_embedding}.
There is only one visual embedding network (VEN) which is shared in both the two tasks. The deep visual embedding (``orange") for a specific item is captured by the trained VEN.}
\label{fig:3d-future-dfsm}
\end{figure*}

Motivated by the observations, we present 3D Furniture shape with TextURE (3D-FUTURE): a richly-annotated, large-scale repository of 3D furniture shapes specific to the household scenario as shown in Figure~\ref{fig:3d-future-overview}. At this time, 3D-FUTURE provides 20,240 realistic indoor images and the associated 9,992 unique 3D furniture models with rich geometry details and informative textures. We render these images via one of the most advanced industrial 3D rendering engines based on 5,000 exquisite room scenes developed by experienced designers. The 3D furniture shapes are used for modern industrial productions and have fine-grained geometry and texture related attributes such as category, style, theme, and material. Further, 3D-FUTURE offers instance segmentation annotation and the rendering information, including six degrees of freedom (6DoF) pose and camera field of view (FoV). Apart from these highlight features, another compelling part of 3D-FUTURE is that it enables many fundamental studies and new research opportunities such as furnishing composition, texture recovery, and other interior understanding subjects.  

It is, however, nontrivial to collect thousands of aesthetic interior designs. To the best of our knowledge, it takes a designer several days to complete a house's interior design. Thus, we considered two main research questions when establishing 3D-FUTURE: 1) can we develop a framework that allows creators to design delicate rooms efficiently? 2) can we automatically create some aesthetic designs based on the professional layout information? To investigate the former question, we build a furnishing suit composition (FSC) platform\footnote{https://3d.shejijia.com/}. The system recurrently recommends visually matched furniture by considering instance aesthetics and compatibility during the design progress. For the latter one, we reuse the expert layouts, generate multiple furnishing suit candidates with some rules and the FSC approach, render the scene, and manually select visually appealing ones. These AI-created designs will also be reviewed by designers to ensure good quality.

The remainder of this paper is organized as follows. First, we briefly review the public 3D benchmarks and discuss their imperfections. Second, we present the data acquisition process and the FSC pipeline. Third, we introduce the properties and statistics of 3D-FUTURE. Finally, we conduct various experiments leveraging on the properties. These experiments can serve as baselines for subsequent research on 3D-FUTURE. 








\begin{figure*}[th!]
\centering
\includegraphics[width=0.98\textwidth]{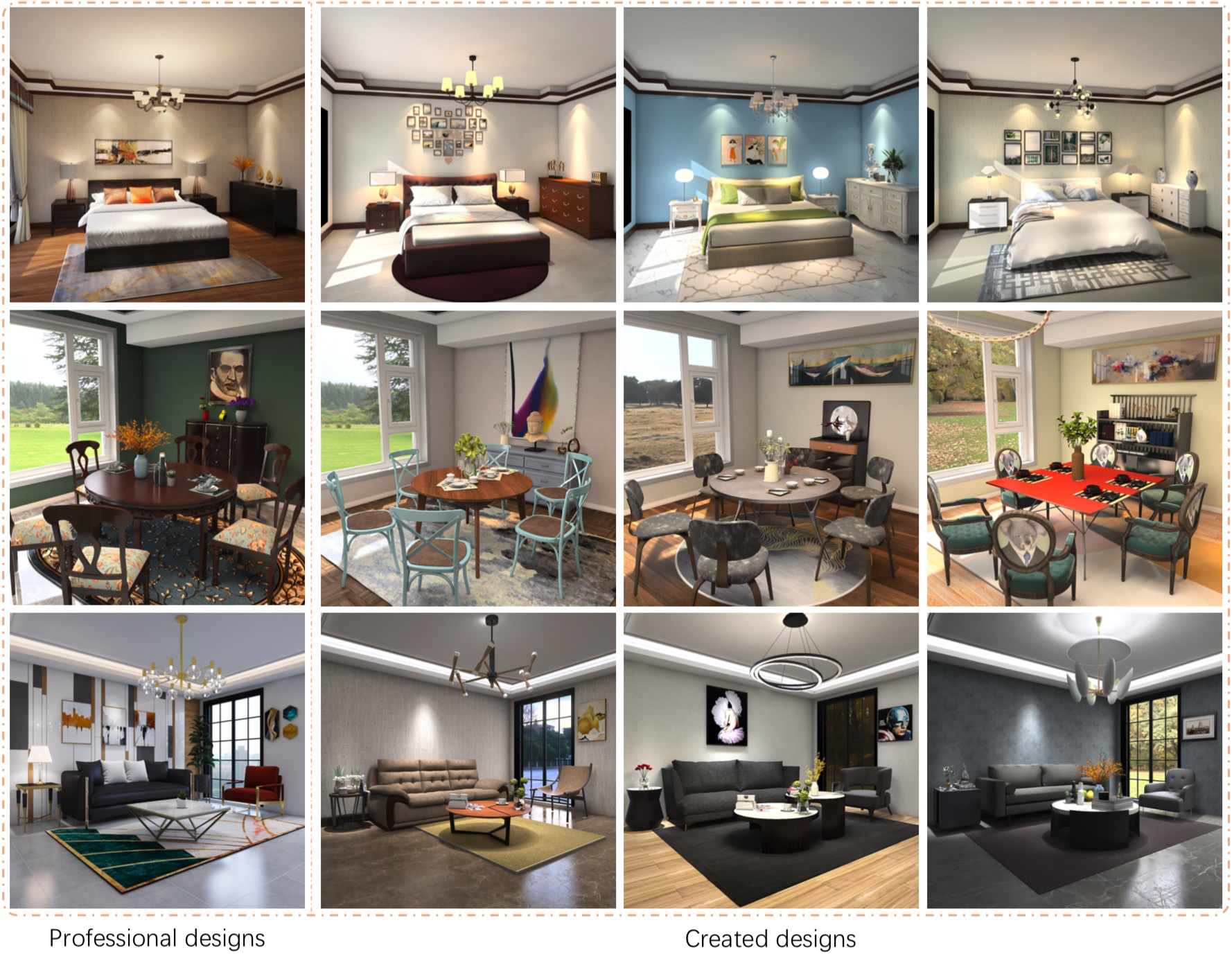}
\caption{\textbf{Realistic Renderings of Aesthetic Interior Designs.} Left: experienced design templates. Right: created aesthetic interior designs. These AI generated designs are reviewed by designers. Zoom in for better view.}
\label{fig:created-design}
\end{figure*}

\section{Related Work}
\label{sec:related_work}
Lots of 3D benchmarks have been established and made publicly available over the past decades \citep{chang2015shapenet,wu20153d,xiao2013sun3d,song2015sun,xiao2016sun,sun2018pix3d,xiang2014beyond,xiang2016objectnet3d,silberman2012indoor,dai2017scannet,hua2016scenenn,choi2016large,shilane2004princeton}. These datasets can be mainly divided into two groups, including 3D models and RGB-D scenes. We will briefly review some representative 3D benchmarks in the following. 

\subsection{3D Models}
\label{subsec:3d_models}
One of the large and exhaustively studied 3D shape repositories is ShapeNet \citep{chang2015shapenet}. It collected millions of raw 3D CAD models from public online repositories such as Warehouse3D and Yobi3D. By re-organizing the datasets, the subsets ShapeNetCore and ShapeNetSem have been made available, including 51,300 and 12,000 models. ShapeNet assigned rich semantic annotations to part of the shapes, such as synsets in the WordNet taxonomy, functional patterns, parts, keypoints, and categories. 3D shape repositories like ModelNet \citep{wu20153d} and Princeton Shape Benchmark \citep{shilane2004princeton} also share similar content as ShapeNet. Several other works like \citep{choi2016large} and ScanObjectNN \citep{uy2019revisiting} create the datasets of 3D scans of real objects based on state of the art (SOTA) RGB-D reconstruction approaches. These benchmarks have largely driven the fundamental 3D studies, including 3D representation, 3D shape recognition, 3D object reconstruction, and part segmentation. However, since the 3D shapes are collected online, many may lack geometry details and have dreamlike or no textures.

Relying on these large-scale 3D shape databases, the community also builds benchmarks with image and shape associations to facilitate the research of 3D object understanding from images. For example, PASCAL3D+ \citep{xiang2014beyond} and ObjectNet3D \citep{xiang2016objectnet3d} aligned objects in the 2D images with the 3D shapes and provided raw 3D pose annotation. Further, Pix3D \citep{sun2018pix3d} contributed more accurate 2D-3D alignment for 395 3D shapes of nine object categories. Unluckily,  these pseudo alignments may largely ignore some local shape details. Moreover, the expensive labor efforts make it difficult to build a large-scale benchmark with precise pixel-level 2D-3D alignment.
\begin{figure*}[th!]
\centering
\includegraphics[width=0.98\textwidth]{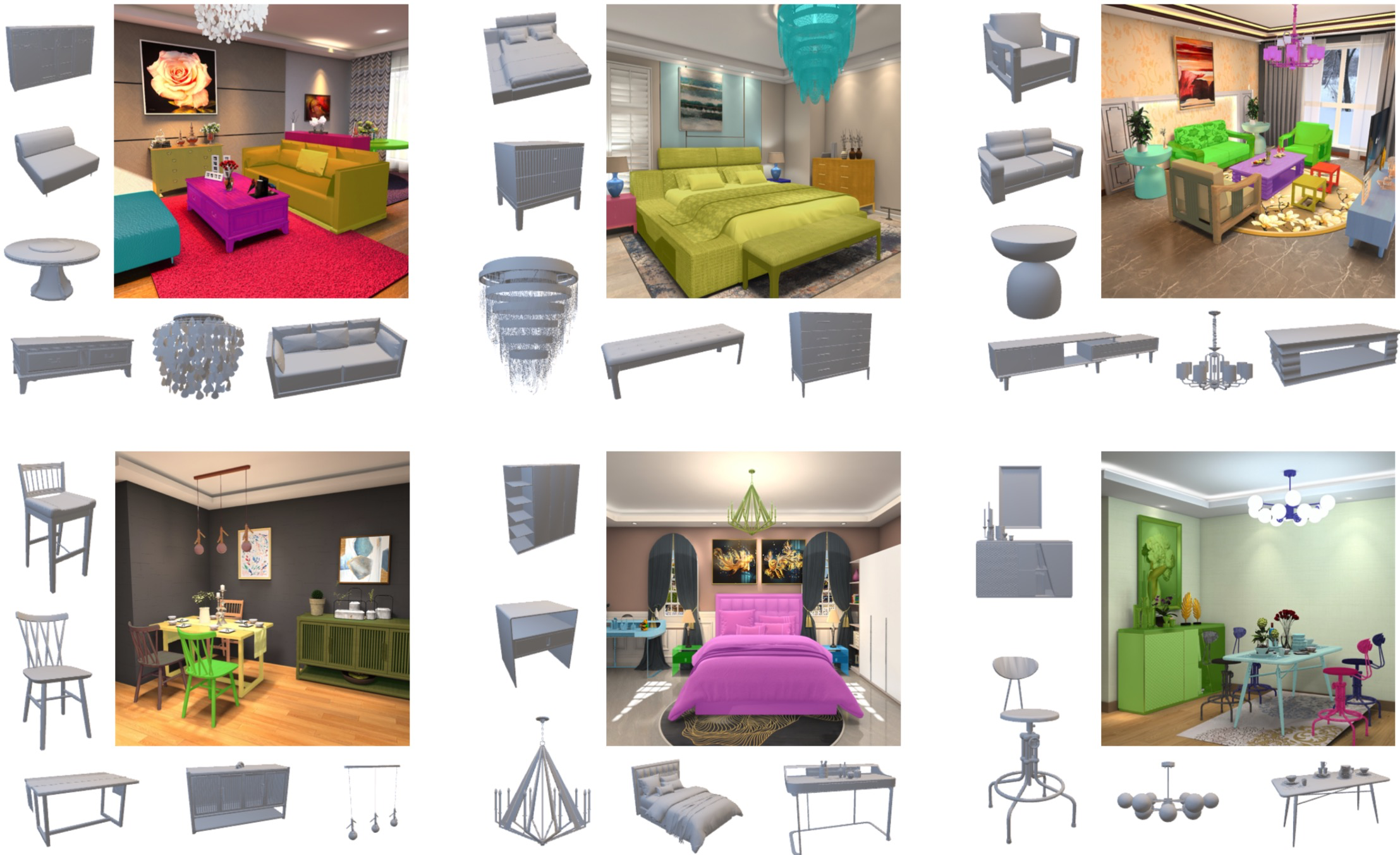}
\caption{\textbf{2D-3D Alignments.} We provide precise 6DoF pose annotations for most of furniture shapes involved in each scene. zoom in for better view.}
\label{fig:real-2d-3d-alignments}
\end{figure*}

\subsection{RGB-D Scenes}
\label{subsec:rgb-d_scenes}
In recent years, the community has put significant efforts into building RGB-D datasets to expand researches on 3D scene understanding. For example, NYU Depth V2 \citep{silberman2012indoor} captured 464 short Kinect RGB-D sequences from 464 different indoor scenes, where 1,449 images are with dense per-pixel labeling, including depth, surface normal, and semantic labels. SUN RGB-D \citep{song2015sun} followed the pattern by annotating 10,335 RGB-D frames, and offered 3D bounding boxes. To capture the full 3D extent of indoor environments, SUN3D \citep{xiao2013sun3d} obtained 415 long sequences in 254 unique spaces with comprehensive views. Further, Dai \emph{et al.} established ScanNet \citep{dai2017scannet}, an RGB-D video dataset containing 2.5M views in 1513 scenes annotated with estimated 3D camera poses, surface reconstructions, semantic segmentation, and a broad set of CAD model alignments. Later, a more extensive dataset Matterport3D \citep{chang2017matterport3d} was made publicly available, contributing to panoramic HDR color images with 3D scene annotations. Different from these RGB-D real-world databases, we focus on experienced exquisite interior designs used in industrial productions.

The works most closely related to ours are InteriorNet \citep{li2018interiornet} and Structured3D \citep{zheng2019structured3d}, which also offer photorealistic images by rendering professional house designs. However, there are two significant differences. First of all, we provide furniture shapes with textures in the scenes. The 6DoF pose and camera FoV are shared in 3D-FUTURE. Second, 3D-FUTURE additionally expects to foster studies of exquisite interior design understanding. Thus, for each room, the camera viewpoints are suggested by designers, so that the captured images contain the whole design idea.
\begin{figure*}[th!]
\centering
\includegraphics[width=0.96\textwidth]{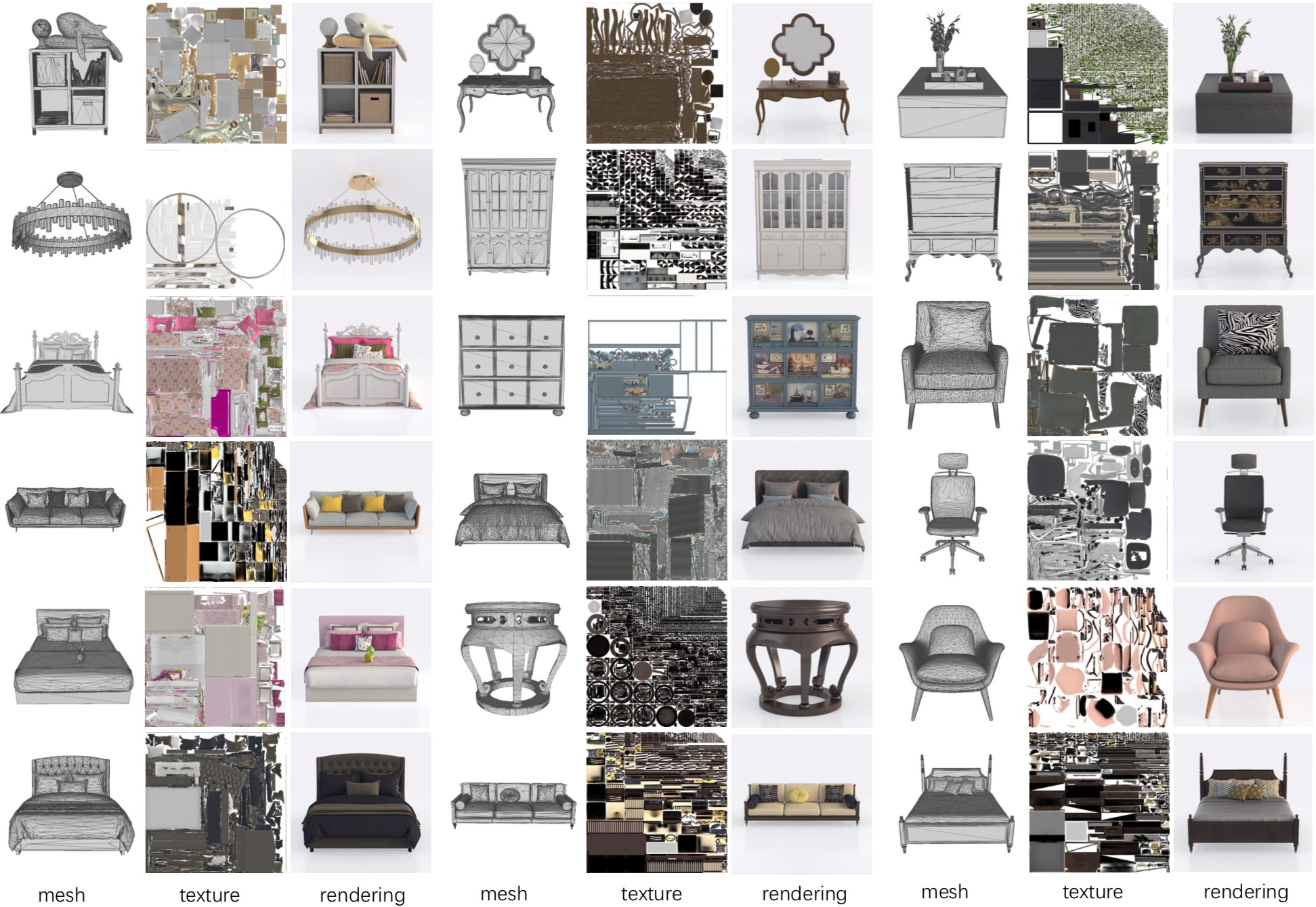}
\caption{Samples of the high-quality 3D shapes and their informative textures in 3D-FUTURE.}
\label{fig:shape-texture}
\end{figure*}

\begin{figure}[t!]
\centering
\includegraphics[width=0.46\textwidth]{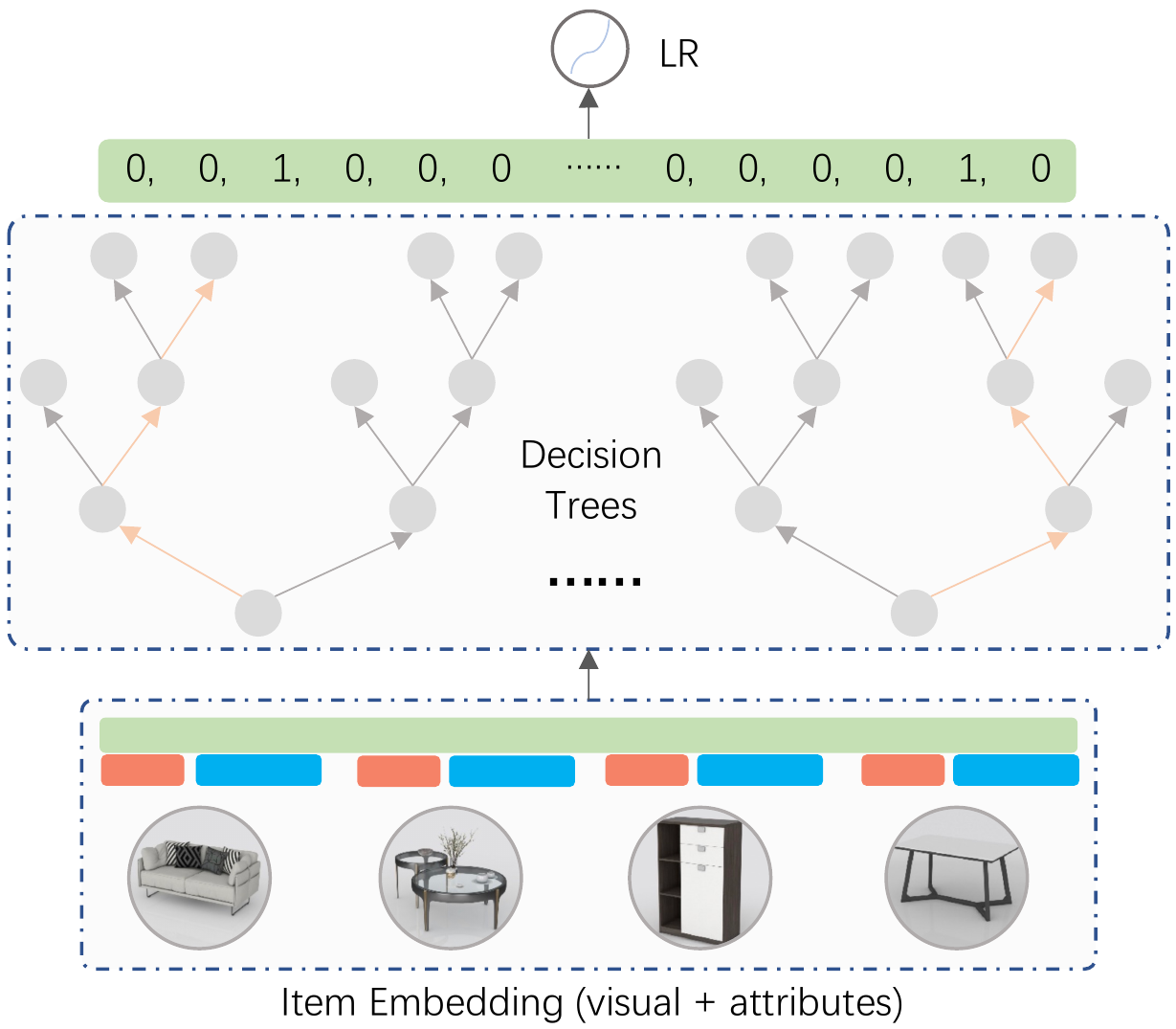}
\caption{An illustration of decision tree based FSC presented in Sec.~\ref{subsubsec:decision_tree}. Orange: The visual embedding for each item is obtained from the trained DFSM in Figure~\ref{fig:3d-future-dfsm}. Blue: The attribute embedding obtained from the attribute labels for each furniture item.}
\label{fig:3d-future-gbdt}
\end{figure}

\section{Data Acquisition Process}
\label{sec:data_acquisition_process}
In this section, we introduce the pipeline of our dataset construction procedure. We mainly address the two issues, \emph{i.e.,} designing efficiency and aesthetic design creation, as stated in Sec.~\ref{sec:intro}.


\subsection{Large-scale Interior Database}
\label{subsec:ms_interior_data}

We construct a 3D pool containing a large amount of industrial 3D computer-aided design (CAD) furnishing and interior finish models. We associate each shape with multiple textures and materials, resulting in enlarged shape repositories. The models are richly annotated with diverse attributes, including theme color, style, material, brand, real-world size, and category in the WordNet taxonomy. There are 500 fine-grained categories in five levels of the taxonomy. High-resolution 2D rendering for each textured model is also available in the database. Based on these objects, hundreds of experienced designers have created $\sim$60,000 decorative houses for different scenarios in several years, where $\sim$30,000 homes have been evaluated as excellent or brilliant designs. A design sample is shown in Figure~\ref{fig:3d-future-overview}. The large-scale interior data is offered by Alibaba Topping Homestyler\footnote{https://www.tangping.com/}. We set up a project based on the large-scale interior data to build 3D-FUTURE.

\subsection{Furnishing Suit Composition (FSC)}
\label{subsec:furnishing_outfit_composition}

One of the main challenges in establishing 3D-FUTURE is how to collect many exquisite interior designs in an acceptable project cycle. To address this issue,  we develop a high-performing furnishing suit composition framework. We mainly borrow the concepts of attribute-based
interpretable compatibility methods \citep{yang2019interpretable,wang2018tem,chen2019pog} in fashion outfit compatibility learning. An overview of this framework is presented in Figure~\ref{fig:3d-future-dfsm}.

Our training set for FSC has the form as $\{\mathcal{X}^1, \mathcal{X}^2, ..., \mathcal{X}^N\}$, and $\mathcal{X}^i = \{x^i_{1}, x^i_{2}, ..., x^i_{m_{i}}\}$. Here, $N$ is the total number of experienced house designs. $m_i$ is the number of items in house $\mathcal{X}_i$, and $x^i_j$ is a specific furnishing item contained in house $\mathcal{X}^i$. Note that the elements in $\mathcal{X}^i$ are in order, which means $x^i_{j}$ is a former furnishing item selected by designers followed by $x^i_{j+1}$. 
\begin{figure*}[th!]
\centering
\includegraphics[width=0.98\textwidth]{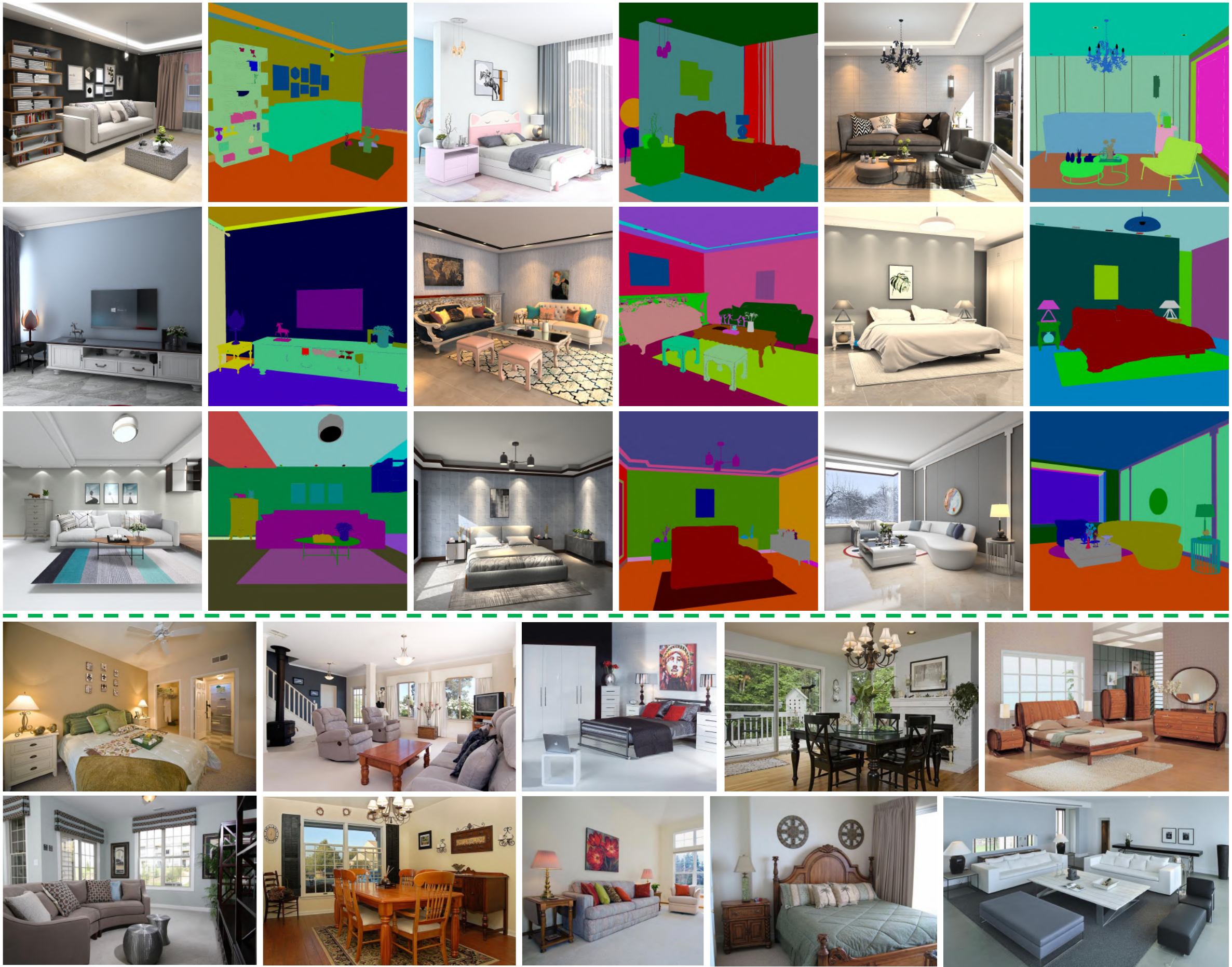}
\caption{Top: samples of photo-realistic synthetic images and their corresponding instance-level annotations from 3D-FUTURE. Bottom: \textbf{\textcolor{red}{natural images}} from the widely studied large-scale scene parsing benchmark ADE20K \citep{zhou2017scene}. Zoom in for better view.}
\label{fig:photorealistic-synthetic-images}
\end{figure*}

\subsubsection{Deep Visual Embedding}
\label{subsubsec:deep_visual_embedding}
As aforementioned, we have rich attribute annotations for each furnishing item. These interpretable attributes show significance in understanding the item's content and are thus beneficial to FSC learning. However, we can not expect a limited number of attributes to represent an item comprehensively. Therefore, we propose a Deep Furnishing Suit Model (DFSM), which consists of a visual embedding network (\emph{VEN}) and two transformer encoders (\emph{TransEnc1} and \emph{TransEnc2}), to learn representative deep visual embedding leveraging on the large-scale excellent house designs. DFSM is driven by a margin ranking loss with hard sample mining and a variant of classification loss.


In specific, given a furnishing suit $\mathcal{X}^i$, we randomly capture a subset $X^{i}_{j \sim k} = \{x^i_j, x^i_{j+1}, ..., x^i_{k}\}$, where $1 \le j \le k < m_i$. Our goal is to predict $x^i_{k+1}$ given $X^{i}_{j \sim k}$. According to the category label of $X^{i}_{j \sim k}$, we randomly choose three negative examples from the furnishing pool to construct a candidate set $C = \{x^i_{k+1}, z^0_{x^i_{k+1}}, z^1_{x^i_{k+1}}, z^2_{x^i_{k+1}}\}$ in an online manner. We also ensure that the negative examples $z^0$ / $z^1$ / $z^2$  have the same style / color / material as $x^i_{k+1}$. We feed both $X^{i}_{j \sim k}$ and the candidate images into \emph{VEN} to extract visual features. In our paper, we take the CNN part of MobileNetV2 followed by a projection layer as \emph{VEN}, and pre-train it via the unsupervised learning strategy stated in \citep{wu2018unsupervised}. 

\begin{figure*}[th]
\centering
\centering
\includegraphics[width=0.98\textwidth]{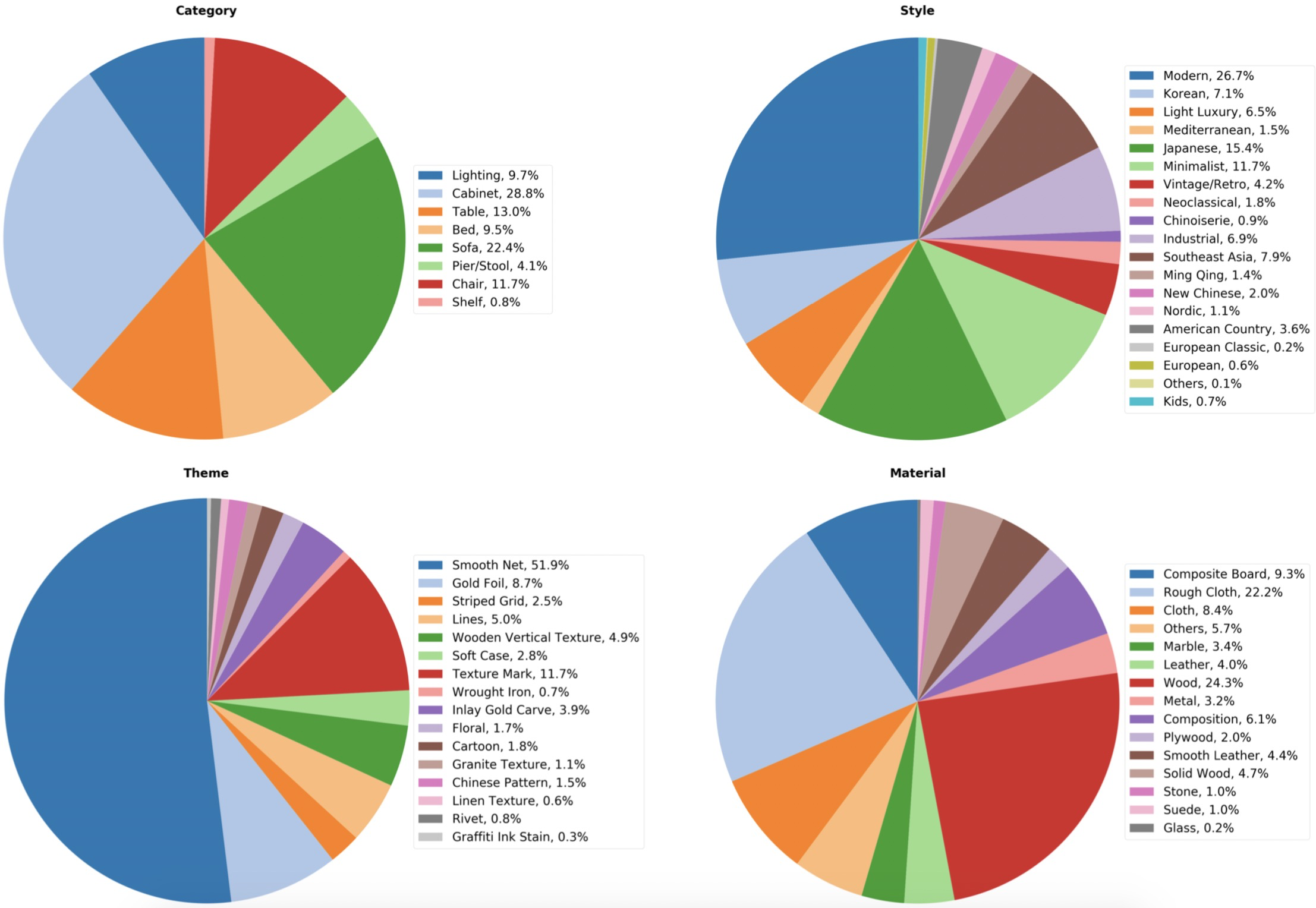}
\caption{The statistics of the attribute annotations of the 9,992 shapes in 3D-FUTURE. Furniture shapes with the attributes such as ``Modern", ``Japanese", ``Smooth Net", ``Texture Mark", ``Rough Cloth", and ``Wood" may be more welcomed by designers when designing the rooms. Besides, except for some special cases, each attribute category has at least 90 shapes.}
\label{fig:attribute}
\end{figure*}

After obtaining the image features, we construct two tasks, \emph{i.e.,} mask prediction and compatibility scoring, based on the impressive transformer architecture in NLP \citep{devlin2018bert,vaswani2017attention}. For the former one, we have a sequence of feature vectors $\mathcal{F}^i = $\{\emph{VEN}($X^{i}_{j \sim k}$), [\emph{Mask}]\} with dimension $d$, where [\emph{Mask}] denotes a particular mask embedding. The task is to predict the masked item given the previous ones. We thus feed $\mathcal{F}$ into \emph{TransEnc1} to capture the enhanced feature $\mathcal{\widetilde{F}}^i$, and optimize the model via the following loss:
\begin{align}
\centering
& \mathcal{L}_{mp} = -\frac{1}{N} \sum_{i=1}^{N}\log(\mathcal{P}(x^i_{k+1}|\mathcal{\widetilde{F}}^i; \Theta, \Phi)), \\
& \mathcal{P}(x^i_{k+1}|\mathcal{\widetilde{F}}^i; \Theta, \Phi)) = \frac{exp(\tilde{f}^i_{mask}f^{T}_{x^i_{k+1}})}{ \sum_{c}exp(\tilde{f}^i_{mask}f^{T}_{c})},
\end{align}
where $\Theta$ and $\Phi$ are the learnable parameters of \emph{VEN} and \emph{TransEnc1}, respectively; $c \in C$ is a candidate; $f^T_{x}$ is the transpose of $f_x$; $f_x$ denotes the visual embedding of item $x$, \emph{i.e.,} $f_x = $ \emph{VEN}($x$); and $\tilde{f}^i_{mask} \in \widetilde{F}^i$ represents the feature vector of the [Mask] token from \emph{TransEnc1}.

For the second task, we take the candidate suits as inputs and directly learn their compatibility scores. Let $F_{(X^{i}_{j \sim k}, c)} = $ \{[Start], \emph{VEN}($X^{i}_{j \sim k}$), $f_c$\} be the visual feature vectors of a candidate suit $O_{(X^{i}_{j \sim k}, c)}$, where [Start] is a particular start token embedding. To estimate the compatibility score of a suit, we need to first capture an embedding that can represent it. We thus employ \emph{TransEnc2} to acquire $\widetilde{F}_{(X^{i}_{j \sim k}, c)}$, and use the feature vector of the [Start] token as the representation of suit $O_{(X^{i}_{j \sim k}, c)}$ (denoted as $r_{(X^{i}_{j \sim k}, c)}$). Further, we utilize two fully connected layers and a sigmoid function to secure a score ($s_{(X^{i}_{j \sim k}, c)}$), which is the measure of the quality of the suit. For conventional presentation, $s_{(X^{i}_{j \sim k}, c)}$ is abbreviated as $s(x^i_{k+1})$ hereafter. Since the ground truth compatibility scores are not available, we minimize a margin ranking loss with a simple hard sample mining policy. The objective is expressed as:
\begin{align}
\centering
& \mathcal{L}_{cs} = -\frac{1}{N} \sum_{i=1}^{N} max(0, -s(x^i_{k+1}) + s(z_{x^i_{k+1}}) + \alpha), \\
& s(z_{x^i_{k+1}}) = max(s(z^0_{x^i_{k+1}}), s(z^1_{x^i_{k+1}}), s(z^2_{x^i_{k+1}})),
\end{align}
where $\alpha$ is set to 0.1 in our experiments. The trained visual embedding network (VEN) is used to extract the visual feature for each furniture item.

\begin{figure*}[th!]
\centering
\centering
\includegraphics[width=0.95\textwidth]{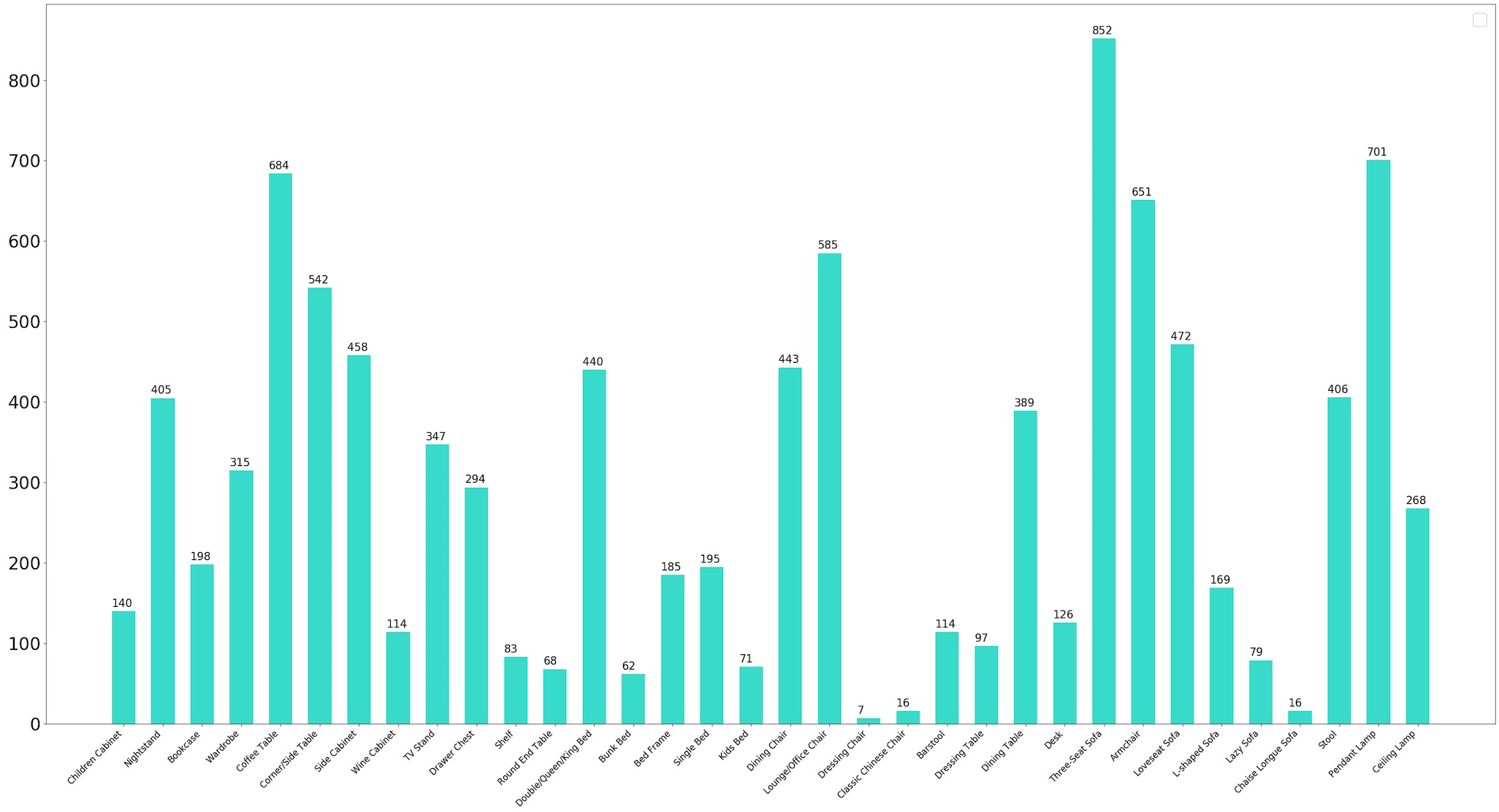}
\caption{The shape number of the 34 categories in 3D-FUTURE. These categories are verified and used by experienced designers in their daily works. The figure also implies the frequency of furniture selected by designers to design the room scenes. There are only 7 dressing chairs because designers commonly choose other chairs as the replacements of dressing chairs when designing a room. For example, Classic Chinese Chair and Chaise Longue Sofa only appear in some special designs.}
\label{fig:shape-hist}
\end{figure*}

\begin{figure*}[th!]
\centering
\centering
\includegraphics[width=0.95\textwidth]{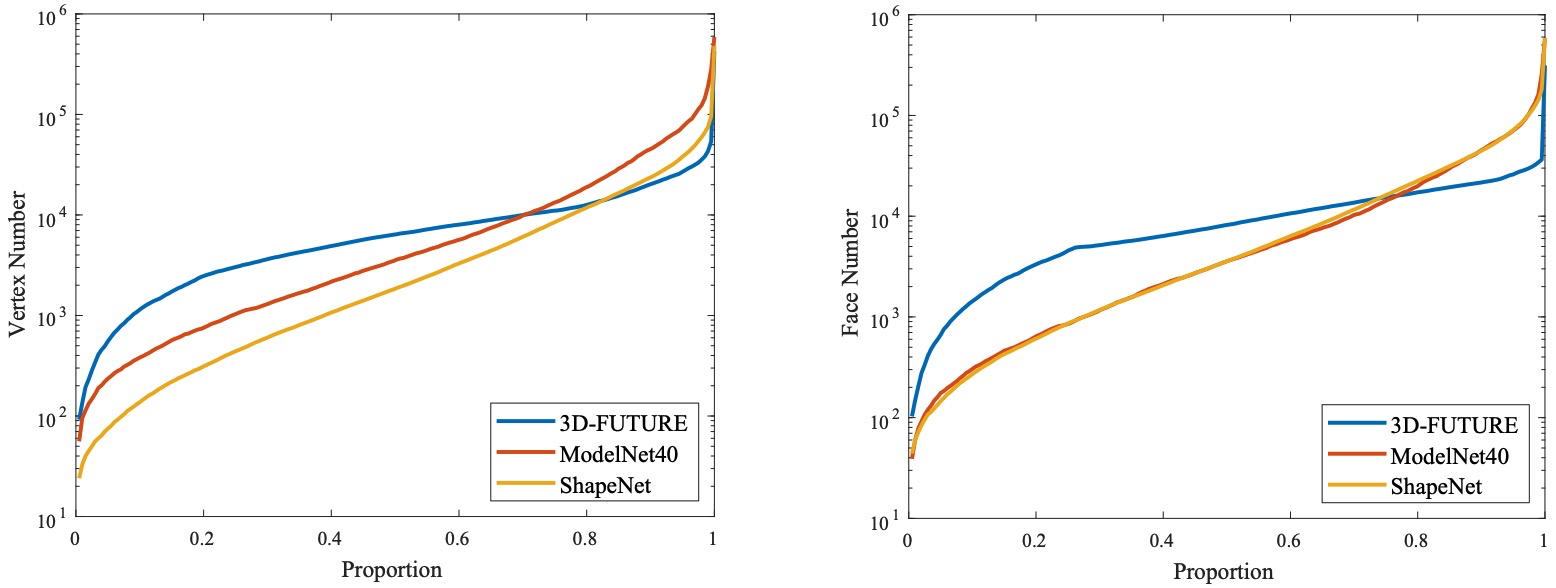}
\caption{The percentile plot of the number of vertices and faces over ShapeNetCore \citep{chang2015shapenet}, ModelNet40 \citep{wu20153d} and 3D-FUTURE. While other datasets have some extremely low-resolution shapes, 3D shapes in 3D-FUTURE show uniformed distributions on both vertices and faces.}
\label{fig:statistic}
\end{figure*}

\subsubsection{Decision Tree Based FSC}
\label{subsubsec:decision_tree}


The main goal here is to infer attribute-based matching patterns, i.e., attribute crosses, for FSC. Considering both interpretability and scalability, we utilize GBDT \citep{friedman2001greedy} to automatically construct attribute crosses as shown in Figure~\ref{fig:3d-future-gbdt}. We will not introduce the details of GBDT here, but only present some facts in training the decision trees. 

We employ six attributes to represent a specific item, including theme color, style, material, real-world size, the second-level category, and visual information (the learned visual embedding). Here, we denote the learned visual embedding as an attribute. For the discrete attributes (style, material, and the second-level category), we directly convert them to one-hot vectors. For theme color and real-world size, we first adopt k-Means Clustering \citep{kanungo2002efficient} to discretize real values and then transform them into one-hot vectors. By further considering the visual embedding, we can represent each item as a feature vector. We assign a label (positive or negative) to each specific furnishing suit to train the decision trees. Both the positive and negative suits are constructed similarly in Sec.~\ref{subsubsec:deep_visual_embedding}.

\begin{figure*}[th!]
\centering
\includegraphics[width=0.96\textwidth]{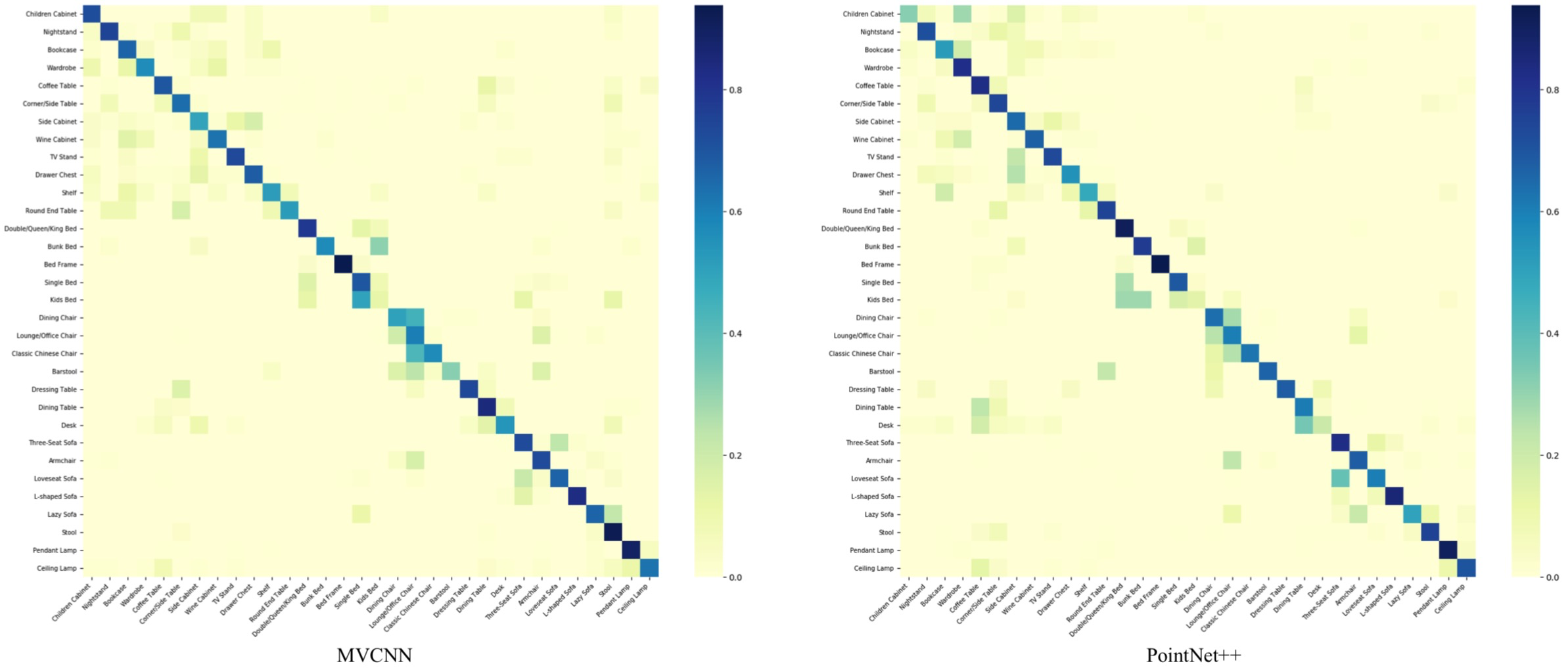}
\caption{The confusion matrices obtained by MVCNN \citep{su2015multi} and PointNet++ \citep{qi2017pointnet++} for 3D Object Recognition on 3D-FUTURE.}
\label{fig:confusion-matirx}
\end{figure*}
\begin{figure*}[th!]
\centering
\includegraphics[width=0.96\textwidth]{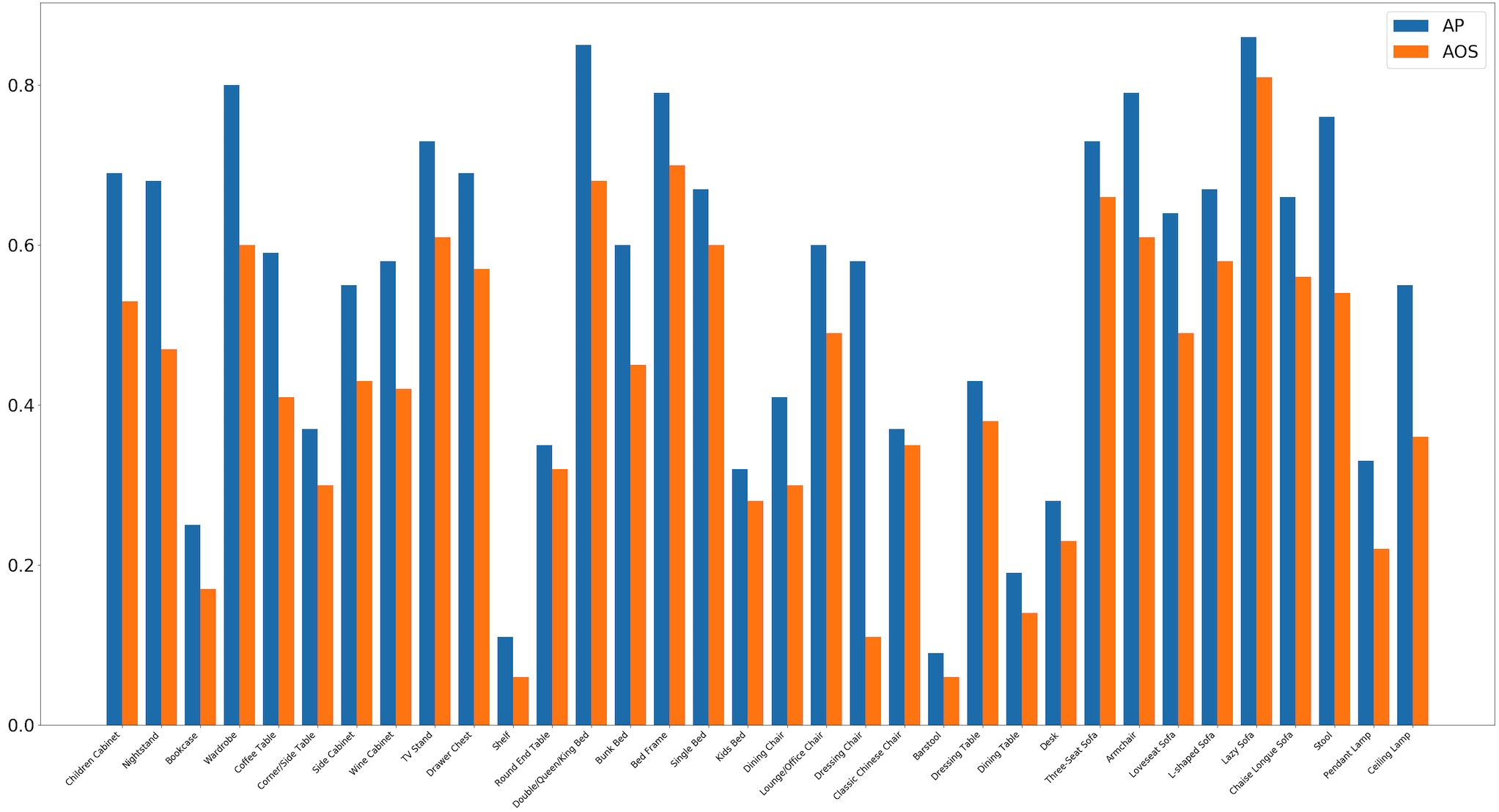}
\caption{Histograms of the instance segmentation AP and rotation estimation AOS of the 34 categories on the test set. The closer AOS is to AP, the better the rotation estimation.}
\label{fig:histogram-aos}
\end{figure*}

\subsubsection{Re-training via Hard Sample Mining}
\label{subsubsec:hard_sample_mining}
The negative furnishing suits construction strategy may return some naive negative samples, due to the large-scale furnishing pool, causing some inaccuracies in both the deep embedding networks and the decision trees. We fine-tune the visual embedding network and re-train the decision tree model via a straightforward hard sample mining strategy to address the issue. Specifically, given $X^{i}_{j \sim k}$, we can have the TopK recommendations using the trained DFSM. We then randomly select negative samples from the TopK pool. After the re-training stage, we fix VEN's parameters and establish an automatically re-training system to update the decision tree model daily using continuously enlarged online designs.

\subsection{Topping Homestyler Design Platform}
\label{subsec:homestyler_design_platform}
Our DFSM is integrated into the online Topping Homestyler Design Platform\footnote{http://3d.shejijia.com} to improve the house design efficiency. There are also other highlight features that can facilitate the design procedure, such as the large-scale shape pool, image-based furniture retrieval, 2D display, 3D Roaming, various professional design templates, texture and item replacement, and online rendering.

\begin{figure*}[th!]
\centering
\includegraphics[width=0.98\textwidth]{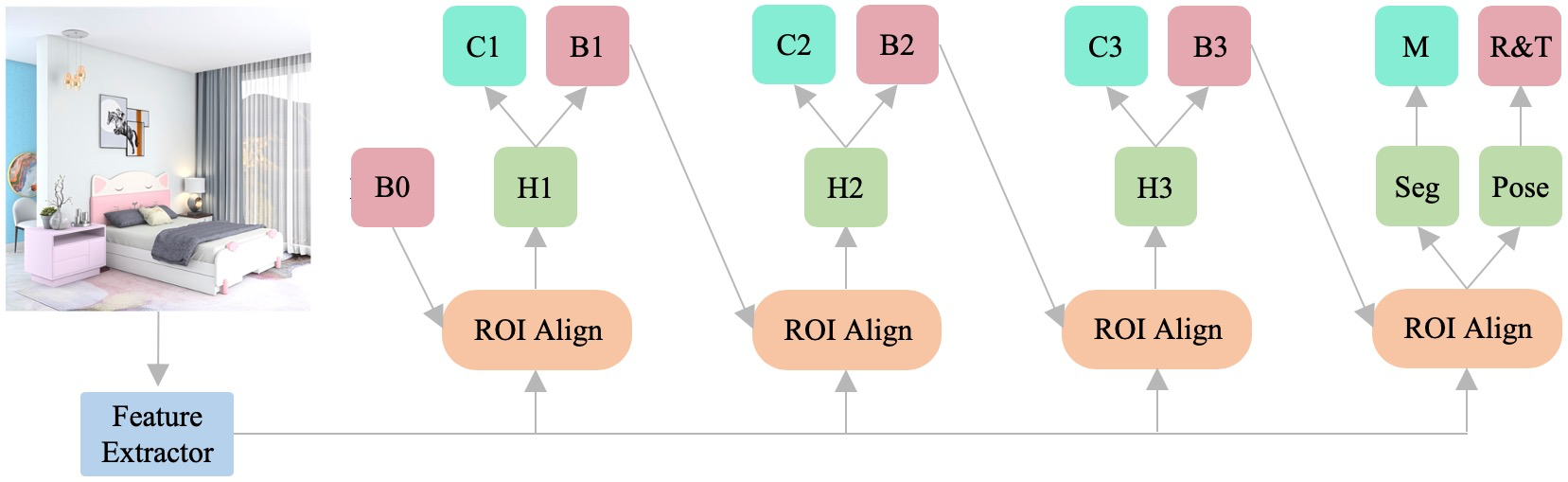}
\caption{An illustration of the network for joint instance segmentation and pose estimation. \textbf{B}: region proposals. \textbf{C}: object recognition. \textbf{H}: network head. \textbf{M}: mask prediction. \textbf{R\&T}: pose estimation. \textbf{Seg}: the network in the instance segmentation branch. \textbf{Pose}: the network in the pose estimation branch.}
\label{fig:instance-net}
\end{figure*}

\begin{table}[h!]
\centering
\begin{tabular}{ c || c  c  }
\hline
Category & MVCNN & PointNet++ \\
\hline
Children Cabinet & 72.0\% & 32.1\% \\
Nightstand & 75.0\% & 71.8\% \\
Bookcase & 66.7\% & 52.3\% \\
Wardrobe & 56.7\%& 82.0\% \\
Coffee Table & 69.7\%& 82.6\% \\
Corner/Side Table & 64.5\% & 74.7\% \\
Side Cabinet & 49.7\% & 65.2\% \\
Wine Cabinet &62.9\% & 67.1\% \\
TV Stand &73.5\% & 73.6\%\\
Drawer Chest & 67.5\% & 55.2\%\\
Shelf &51.9\% & 48.4\%\\
Round End Table &52.2\% & 75.0\%\\
Double/Queen/King Bed &78.6\% & 91.2\%\\
Bunk Bed &57.1\% & 77.8\%\\
Bed Frame &93.8\% & 93.8\%\\
Single Bed &69.7\% & 68.9\%\\
Kids Bed &12.5\% & 14.3\%\\
Dining Chair &50.5\% & 63.9\%\\
Lounge/Office Chair &60.3\% & 60.5\%\\
Classic Chinese Chair &57.1\% & 62.5\%\\
Barstool &32.0\% & 66.7\%\\
Dressing Table &73.7\% & 68.2\%\\
Dining Table &84.3\% & 61.1\%\\
Desk &54.0\% & 20.4\%\\
Three-Seat Sofa &71.7\% & 82.6\%\\
Armchair &72.5\% &  68.0\%\\
Loveseat Sofa &62.9\% & 60.4\%\\
L-shaped Sofa & 83.3\%& 85.9\%\\
Lazy Sofa &66.7\% & 50.0\%\\
Stool & 91.9\%& 75.8\%\\
Pendant Lamp &89.8\% & 90.9\%\\
Ceiling Lamp & 63.0\%& 70.7\%\\
\hline
mean & 69.2\% & 69.9\%\\
\hline
\end{tabular}
\caption{Classification accuracy on 3D-FUTURE. MVCNN: 12 view + ResNet50 backbone. PointNet++: 1024 points + MSG + normal.}
\label{tab:recognition}
\end{table}

\begin{figure}[h!]
\centering
\includegraphics[width=0.48\textwidth]{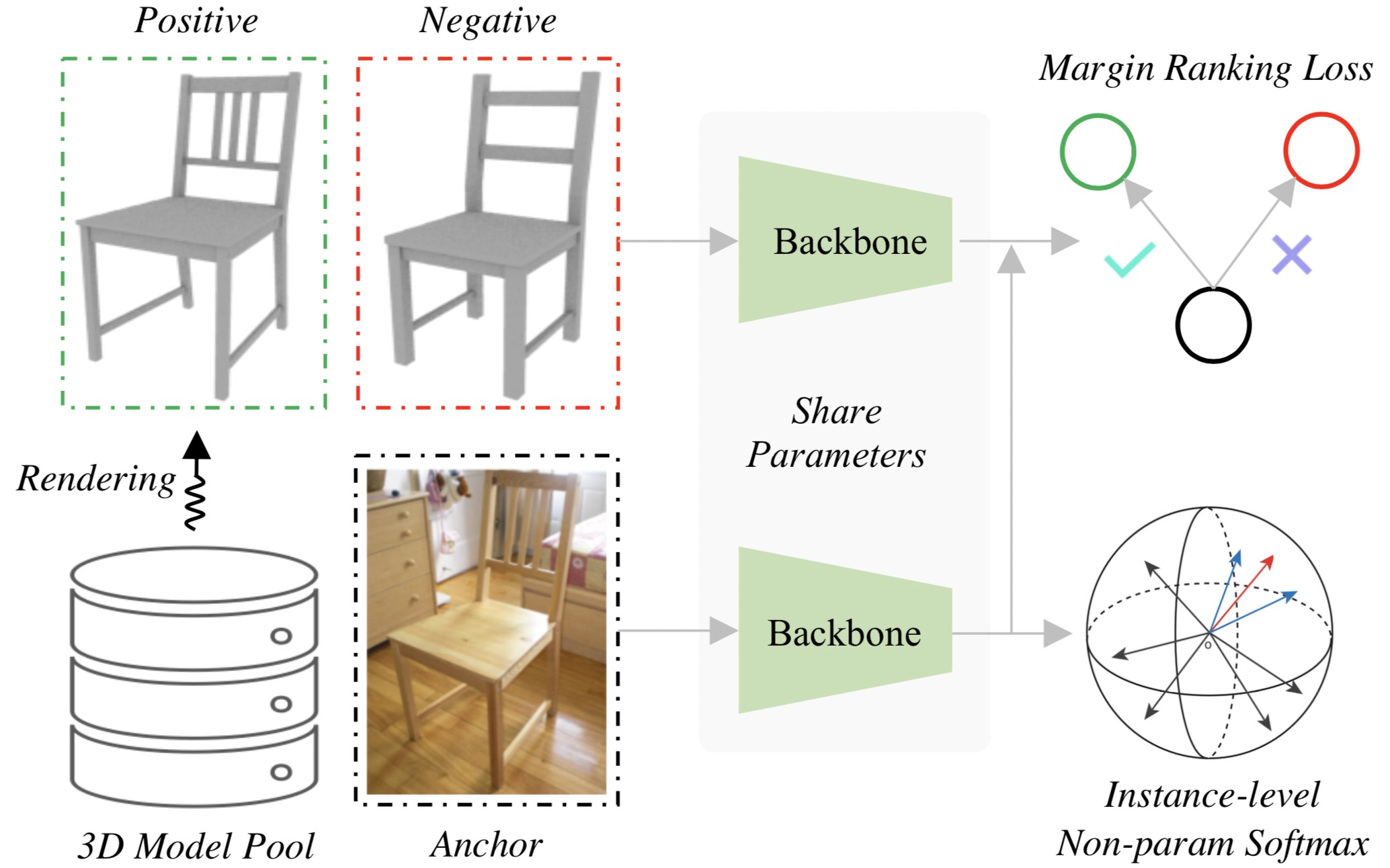}
\caption{An illustration of the baseline method of cross-domain image-based 3D shape retrieval. We use instance-level non-parametric softmax loss \citep{wu2018unsupervised} so that the network can capture shape similarity among furniture instances.}
\label{fig:retrieval-net}
\end{figure}

\subsection{Create Aesthetic Interior Design}
\label{subsec:create_aesthetic_interior_design}
We have collected 5,000 exquisite interior room designs in the 3D-FUTURE project. We do not plan to provide several synthetic images in different viewpoints for each room. Instead, we expect to deliver more superlative designs to bring more research possibilities. Thus, we take these experienced designs as templates and create several aesthetic interior designs for each template. 
 
 \begin{figure*}[th!]
\centering
\includegraphics[width=0.96\textwidth]{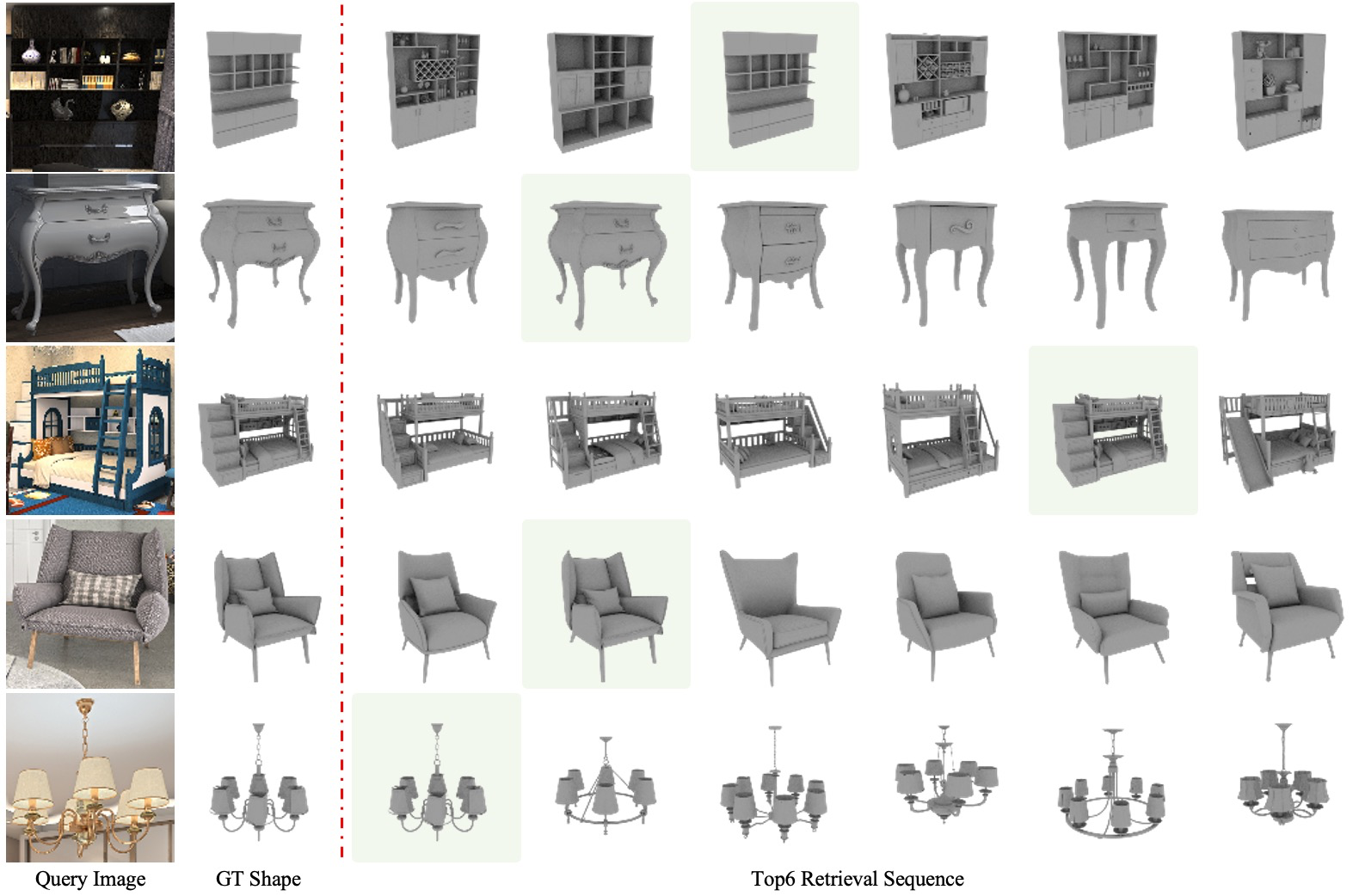}
\caption{The retrieval sequences for several query images. 3D-FUTURE contains fine-grained shapes for each furniture category.}
\label{fig:retrieval-results}
\end{figure*}

\begin{table}[h!]
\centering
\begin{tabular}{ c || c  c | c }
\hline
 & train & test & Occ. Ratio \\
\hline
NO & 17,638 & 3,506 & $<$ 0.1\\
Slight & 8,637 & 1,545 & 0.1 $\sim$ 0.2 \\
Standard & 5,169 & 943 & 0.2 $\sim$ 0.3\\
\hline
Total Image & 31,444 & 5,997 & - \\
\hline
CAD Shape & 6,699 & 3,293 + 6,699 & - \\
\hline
\end{tabular}
\caption{The train and test sets for the subject of cross-domain image-based 3D shape retrieval. Object labeled as ``NO" means the occluded ratio for the object is less than 10\%. In our setting, the final retrieval pool consists of the CAD models from both the test set and the train set.}
\label{tab:retrieval-sta}
\end{table}


For example, given a template room with professional design information, we first replace the interior finishing according to the materials, room style, and other descriptions. Second, we choose a furniture seed (\emph{e.g.}, bed) based on the interior finishing information. Third, we iteratively perform recommendations based on our DFSM and other rules to generate a furnishing list. Finally, we put the items contained in the furnishing list into their corresponding positions. In the third step, we also learn one-to-one visual compatibility models (\emph{e.g,} bed-nightstand and sofa-coffee table) as additional rules to improve the robustness.

With the pipeline, we can automatically create many interior designs as shown in Figure~\ref{fig:created-design}. To ensure the quality, we render an image for each design and manually select 15,240 visually appealing designs. Our experienced designers further review these designs to assure good quality. 

The 3D designs are rendered by one of the most advanced computer-generated imagery rendering software applications, V-Ray\footnote{https://www.chaosgroup.com/}. To ensure reality, we enable as many functions as possible supported by V-Ray.


\section{Properties of 3D-FUTURE}
\label{sec:properties_statistics}
In this section, we summarize the properties of our 3D-FUTURE database. Compared to previous 3D benchmarks, 3D-FUTURE has some prominent properties that can bring more possibilities for future 3D research.

\subsection{Photo-realistic Synthetic Images}

3D-FUTURE offers 20,240 photo-realistic synthetic images corresponding to 20,240 interior designs. As aforementioned, we have 5,000 experienced designs and 15,240 automatically created aesthetic designs. We render one image for each design. Previous datasets, such as Structured3D \citep{zheng2019structured3d} and InteriorNet \citep{li2018interiornet}, also provide realistic indoor images and scene parsing annotations. However, they put cameras in random positions and capture redundant images for each house. These images were not manually verified, thus suffer from unexpected viewpoints.

\begin{table}[th!]
\centering
\begin{tabular}{ c || c  c  | c  }
\hline
Category & Top1@R & Top3@R & F-score \\
\hline
Children Cabinet & 26.9 & 53.4 & 29.7  \\
Nightstand & 37.4 & 64.1 & 32.6 \\
Bookcase & 8.7  & 8.7 & 23.9  \\
Wardrobe & 21.7 & 43.3 & 42.2  \\
Coffee Table & 27.7 & 57.1 & 31.3 \\
Corner/Side Table & 32.0 & 49.8 & 32.4  \\
Wine Cabinet & 4.9 & 7.3 &  22.1 \\
TV Stand & 16.4 & 27.5 & 37.1  \\
Drawer Chest & 15.0 & 31.8 & 44.6 \\
Shelf & 19.4 & 27.3 & 35.6 \\
Round End Table & 20.0 & 21.1 & 26.1 \\
Double/Queen/King Bed & 23.8 & 48.7 & 78.8  \\
Bunk Bed & 13.0 & 26.1 & 50.0  \\
Bed Frame & 26.0 & 47.1 & 52.6  \\
Single Bed & 16.5 & 34.7 & 63.2  \\
Kids Bed & 18.2 & 45.5 & 65.5  \\
Dining Chair & 16.1 & 38.4 & 50.5  \\
Lounge/Office Chair & 33.7 & 64.7 & 56.4 \\
Classic Chinese Chair & 20.0 & 80.0 & 49.8  \\
Barstool & 52.8 & 58.3 & 22.1  \\
Dressing Table & 53.9 & 65.4 & 31.8  \\
Dining Table & 13.9 & 21.5 & 26.0  \\
Desk & 13.2 & 27.9 & 18.6  \\
Three-Seat Sofa & 5.6 & 10.5 & 59.8 \\
Armchair & 45.6 & 68.7 & 56.9  \\
Loveseat Sofa & 6.7 & 17.5 & 56.1 \\
Lazy Sofa & 19.4 & 48.4 & 51.7  \\
Chaise Longue Sofa & 16.7 & 16.7 & 31.4 \\
Stool & 31.6 & 63.3 & 48.9 \\
Pendant Lamp & 29.4 & 56.4 & 48.8 \\
Ceiling Lamp & 31.4 & 59.1 & 37.6 \\
\hline
mean & 23.4 & 40.6 & 47.1 \\
\hline
\end{tabular}
\caption{Numerical retrieval results on 3D-FUTURE for category level. We train a single model and perform retrieval in the full 3D-FUTURE pool. F-score here represents Top5 average F-score.}
\label{tab:retrieval}
\end{table}

In contrast, 3D-FUTURE focuses more on inspiring the understanding of exquisite interior designs. Thus, the camera positions are suggested by professional designers to obtain the best viewpoint for each room. Besides, 3D-FUTURE provides instance semantic labels of 34 categories and ten supper-categories. Moreover, the images contained in 3D-FUTURE are visually more appealing and realistic compared to previous ones.

\subsection{2D-3D Alignments}

Previous benchmarks only provide pseudo 2D-3D alignment annotations \citep{xiang2016objectnet3d,sun2018pix3d,dai2017scannet,KrauseStarkDengFei-Fei_3DRR2013,xiang2014beyond}. Namely, they manually choose a roughly matched 3D CAD model from public 3D shape benchmarks according to the object contained in an image. Annotators thus may largely ignore some local shape details. As a result, these benchmarks offer a small number of matched 3D shape and 2D image pairs. Besides, previous benchmarks with alignment annotations do not come with scene images. In contrast, 3D-FUTURE provides precious 2D-3D alignments and 3D pose annotations. It contains 9,992 unique 3D shapes and 20,240 scene images. By cropping instances from the scene images, we can further secure 37,441 image and shape pairs with slight occlusions, as reported in Table~\ref{tab:retrieval-sta}. Some samples are presented in Figure~\ref{fig:real-2d-3d-alignments}. 

\subsection{High-quality Shapes with Informative Textures}

The 3D shapes contained in previous large-scale shape repositories \citep{chang2015shapenet,shilane2004princeton,wu20153d} are mainly collected from online repositories. These 3D CAD models usually contain few geometry details and low informative textures. Luckily, 3D-FUTURE provides high-quality 3D furniture shapes with rich details in various styles, including European furniture, which often contains intricate carvings. All the shapes come with informative textures and have been used for modern industrial productions. We show some samples in Figure~\ref{fig:shape-texture}. We believe these features can potentially facilitate innovative research on high-quality 3D shape understanding and generation. In 
Figure \ref{fig:statistic}, we compare the proportion of different number of vertices and faces over ShapeNetCore \citep{chang2015shapenet}, ModelNet40 \citep{wu20153d} and our dataset. While other datasets have some extremely low-resolution shapes, 3D shapes in 3D-FUTURE show uniform distributions in both vertices and faces.

\setlength\tabcolsep{7pt}
\begin{table*}[h!]
\centering
\begin{tabular}{ c || c  c | c  c | c  c || c  c | c  }
\hline
Category & AP & AR & AP$^{50}$ & AR$^{50}$ & AP$^{75}$ & AR$^{75}$ & AOS & AVP & RMSE\\
\hline
Children Cabinet & 0.69 & 0.75 & 0.83 & 0.86 & 0.79 & 0.83 & 0.53 & 0.54 & 0.38 \\
Nightstand & 0.68 & 0.75 & 0.94 & 0.97 & 0.81 & 0.86 & 0.47 & 0.48 & 0.83 \\
Bookcase & 0.25 & 0.41 & 0.52 & 0.68 & 0.21 & 0.43 & 0.17 & 0.17	& 0.93 \\
Wardrobe & 0.80 & 0.86 & 0.93 & 0.96 & 0.90 & 0.94 & 0.60 & 0.60 & 0.43  \\
Coffee Table & 0.59 & 0.67 & 0.94 & 0.96 & 0.64 & 0.75 & 0.41 & 0.40 & 0.26  \\
Corner/Side Table & 0.37 & 0.49 & 0.74 & 0.82 & 0.32 & 0.49 & 0.30 & 0.29 & 0.29  \\
Side Cabinet & 0.55 & 0.65 & 0.81 & 0.88 & 0.60 & 0.71 & 0.43 & 0.43 & 0.28  \\
Wine Cabinet & 0.58 & 0.65 & 0.86 & 0.90 & 0.65 & 0.74 & 0.42 & 0.42 & 0.52  \\
TV Stand & 0.73 & 0.79 & 0.95 & 0.96 & 0.89 & 0.91 & 0.61 & 0.62 & 0.38  \\
Drawer Chest & 0.69 & 0.77 & 0.83 & 0.89 & 0.79 & 0.86 & 0.57 & 0.58 & 0.45 \\
Shelf & 0.11 & 0.35 & 0.25 & 0.55 & 0.03 & 0.17 & 0.06 & 0.06 & 0.09 \\
Round End Table & 0.35 & 0.43 & 0.80 & 0.84 & 0.20 & 0.37 & 0.32 & 0.32 & 0.22 \\
Double/Queen/King Bed & 0.85 & 0.92 & 0.95 & 0.98 & 0.91 & 0.96 & 0.68 & 0.69 & 0.24 \\
Bunk Bed & 0.60 & 0.68 & 0.85 & 0.93 & 0.78 & 0.88 & 0.45 & 0.46 & 0.21 \\
Bed Frame & 0.79 & 0.87 & 0.94 & 0.99 & 0.89 & 0.94 & 0.70 & 0.71 & 0.20 \\
Single Bed & 0.67 & 0.79 & 0.79 & 0.88 & 0.73 & 0.84 & 0.60 & 0.61 & 0.28 \\
Kids Bed & 0.32 & 0.59 & 0.47 & 0.76 & 0.36 & 0.65 & 0.28 & 0.28 & 0.33 \\
Dining Chair & 0.41 & 0.50 & 0.79 & 0.84 & 0.37 & 0.53 & 0.30 & 0.29 & 0.14  \\
Lounge/Office Chair & 0.60 & 0.72 & 0.85 & 0.93 & 0.71 & 0.83 & 0.49 & 0.49 & 0.43  \\
Dressing Chair & 0.58 & 0.65 & 0.92 & 0.94 & 0.68 & 0.78 & 0.11 & 0.11 & 0.07  \\
Classic Chinese Chair & 0.37 & 0.40 & 0.77 & 0.81 & 0.28 & 0.33 & 0.35 & 0.36 & 0.07  \\
Barstool & 0.09 & 0.19 & 0.35 & 0.51 & 0.02 & 0.11 & 0.06 & 0.05 & 0.05 \\
Dressing Table & 0.43 & 0.50 & 0.91 & 0.93 & 0.30 & 0.50 & 0.38 & 0.38 & 0.22 \\
Dining Table & 0.19 & 0.32 & 0.63 & 0.76 & 0.06 & 0.22 & 0.14 & 0.13 & 0.11  \\
Desk & 0.28 & 0.43 & 0.72 & 0.82 & 0.15 & 0.35 & 0.23 & 0.23 & 0.15 \\
Three-Seat Sofa & 0.73 & 0.84 & 0.90 & 0.98 & 0.88 & 0.96 & 0.66 & 0.66 & 0.28 \\
Armchair & 0.79 & 0.86 & 0.91 & 0.95 & 0.89 & 0.93 & 0.61 & 0.60 & 0.26  \\
Loveseat Sofa & 0.64 & 0.79 & 0.77 & 0.91 & 0.73 & 0.87 & 0.49 & 0.48 & 0.35 \\
L-shaped Sofa & 0.67 & 0.79 & 0.80 & 0.91 & 0.79 & 0.90 & 0.58 & 0.59 & 0.59 \\
Lazy Sofa & 0.86 & 0.88 & 0.90 & 0.91 & 0.89 & 0.91 & 0.81 & 0.82 & 0.18 \\
Chaise Longue Sofa & 0.66 & 0.76 & 0.84 & 0.91 & 0.84 & 0.91 & 0.56 & 0.52 & 0.42 \\
Stool & 0.76 & 0.82 & 0.90 & 0.93 & 0.85 & 0.89 & 0.54 & 0.52 & 0.22 \\
Pendant Lamp & 0.33 & 0.47 & 0.70 & 0.79 & 0.28 & 0.47 & 0.22 & 0.21 & 0.21  \\
Ceiling Lamp & 0.55 & 0.65 & 0.84 & 0.87 & 0.63 & 0.73 & 0.36 & 0.35 & 0.28 \\
\hline
mean & 0.55 & 0.65 & 0.79 & 0.87 & 0.58 & 0.69 & 0.43 & 0.43 & 0.30 \\
\hline
\end{tabular}
\caption{Quantitative results of the Cascade-Mask R-CNN baseline for joint instance segmentation and 3D pose estimation. AOS and AVP: Higher is better. RMSE: Lower is better.}
\label{tab:pose-estimation}
\end{table*}

\subsection{Fine-Grained Attributes}
Previous 3D benchmark provides functional attribute annotations in WordNet taxonomy for 3D shapes \citep{chang2015shapenet}. However, these attributes are not well organized and do not have corresponding textures. In contrast, for each textured shape in 3D-FUTURE, we provide four types of attributes verified by professional designers. We have 34 shape categories, 8 super-categories, 19 styles, 15 materials, and 16 themes. These attributes have been demonstrated valuable for interior designs and content understanding by industrial productions. We present the statistics of these attributes in Figure~\ref{fig:shape-hist} and Figure~\ref{fig:attribute}. These figures imply the preferences of experienced modern designers when designing the rooms.


\begin{figure*}[th!]
\centering
\includegraphics[width=0.98\textwidth]{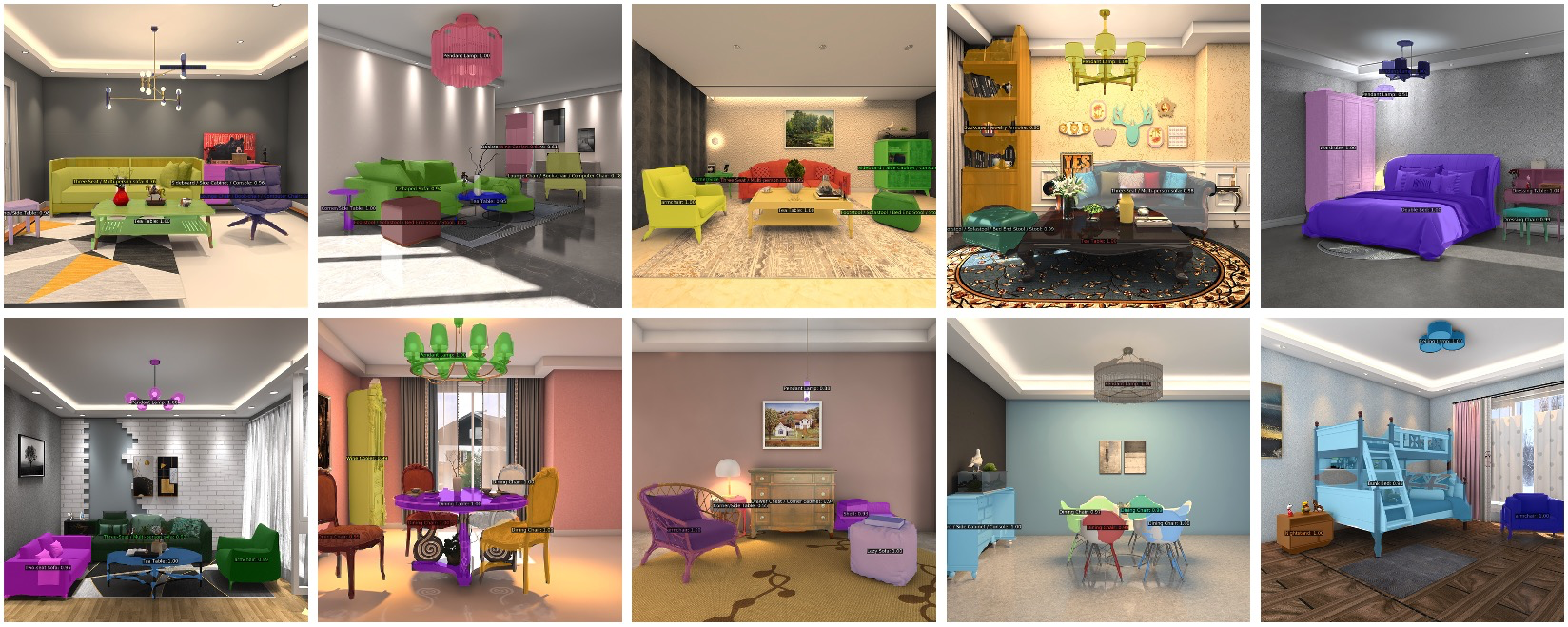}
\caption{Instance segmentation results. The images are captured under suggested viewpoint (by designer) for design exhibition. Zoom in for better view.}
\label{fig:instance-seg}
\end{figure*}

\section{Baseline Experiments}
\label{sec:baseline_experiments}
In the section, we conduct several baseline experiments by leveraging the properties of 3D-FUTURE, including shape recognition, joint 2D instance segmentation and 3D pose estimation, image-based shape retrieval, 3D object reconstruction, and texture synthesis.
We split our 3D shapes into a training set with 6,699 models, and a test set with 3,293 models. The scene images are divided according to the training and test splits of 3D shapes. There are 14,761 images for training and 5,479 images for test. We will briefly present the experimental details for each task and report the scores.
\subsection{Fine-grained 3D Object Recognition}
Over the past years, most 3D object recognition methods extend deep convolutional neural networks (DCNNs) to modeling 3D data. Because 3D CNNs are too memory intensive \citep{ji20123d}, some researchers prefer to either develop special deep learning operations on point clouds and mesh surfaces \citep{qi2017pointnet,qi2017pointnet++,hanocka2019meshcnn,feng2019meshnet}, or project 3D shapes to several 2D images and then apply 2D convolutional networks \citep{su2015multi}. However,  it is nontrivial to extend the projection-based methods to high-resolution 3D scene understanding. Moreover, point and mesh-based approaches suffer from computation bottlenecks and are thus limited to sparse point clouds and a small number of surfaces. 

In contrast to ShapeNet \citep{chang2015shapenet} and ModelNet \citep{wu20153d}, 3D-FUTURE enables the study of fine-grained 3D furniture recognition, which requires the networks to capture more local and global geometric details. Here we consider the well-known MVCNN \citep{su2015multi} and PointNet++ \citep{qi2017pointnet++} as the baselines. In specific, we train a 12-view MVCNN with ResNet50 as the backbone. For PointNet++, we sample 1024 points for each shape instance and adopt the multi-scale grouping (MSG) strategy \citep{qi2017pointnet++} and normal vectors to secure the best performance. We train the networks using 6,699 shapes and evaluate the trained models via the remaining 3293 shapes. The classification accuracy for each category is presented in Table~\ref{tab:recognition} and Figure~\ref{fig:confusion-matirx}. While these methods can reach 90\% accuracy on ModelNet40 and ShapenetCore, they do not perform well (69.2\% $\sim$ 69.9\%) on 3D-FUTURE, due to the presence of fine-grained furniture categories. This observation would motivate researchers to exploit more efficient 3D representation learning approaches for deeper 3D shape analysis.



\subsection{Image-based 3D Shape Retrieval}

Cross-domain image-based 3D shape retrieval (IBSR) is to identify the  CAD models of the objects contained in query images. The primary issue in IBSR is the large appearance gaps between 3D shapes and 2D images. To tackle this challenge, early works made efforts to map cross-domain representations into a unified constrained embedding space via adaptation techniques such as weight-sharing constraints, metric learning, and distance matching \citep{li2015joint,aubry2014seeing,lee2018cross,massa2016deep,tasse2016shape2vec,girdhar2016learning}. 
Recent works \citep{sun2018pix3d,huang2018holistic,wu2017marrnet,bansal2016marr,bachman1978data,grabner20183d,grabner2019location} predict 2.5D sketches from images, such as surface normal, depth, and location field, to bridge the gaps between 3D and 2D domains. However, the performance of state-of-the-art IBSR methods show a large gap than its 2D counterpart, \emph{i.e.,} content-based image retrieval. This is because there are no large-scale benchmarks that offer large amounts of precious 2D-3D alignment annotations.

\setlength\tabcolsep{5pt}
\begin{table*}[th!]
\centering
\begin{tabular}{ c || c  c  c | c  c  c | c  c  c }
\hline
\multirow{2}{*}{Category} & \multicolumn{3}{c|}{IoU (\%)} & \multicolumn{3}{c|}{CD ($\times 10^{-3}$)} & \multicolumn{3}{c}{F-score (\%)} \\ \cline{2-10}
& Pixel2Mesh & ONet & DISN & Pixel2Mesh & ONet & DISN & Pixel2Mesh & ONet & DISN \\
\hline
Children Cabinet & 65.54 & 35.96 & 67.50 & 64.00 &172.36 & 100.56 & 43.19 & 13.76 & 34.96\\
Nightstand & 53.42 & 40.75 & 60.10 & 69.95 & 177.66 & 134.67 & 41.01 & 18.60 & 30.86 \\
Bookcase & 52.74 & 18.15 & 50.89 & 33.04 & 132.13 & 46.14 &59.03 &15.76 & 52.06 \\
Wardrobe & 64.65 &34.10 & 66.63 & 47.69 & 147.30 & 79.37 & 46.19 & 15.20 & 38.71 \\
Coffee Table & 42.74 & 15.77 & 41.89 & 58.55 & 165.69 &92.69 & 49.24 & 17.24 & 38.93 \\
Corner/Side Table & 38.08 & 17.55 & 42.37 &51.02 & 213.03 & 123.63 & 79.26 & 17.78 & 35.00 \\
Side Cabinet & 60.10 & 30.47 & 66.31 & 44.34 & 121.10 & 60.33 & 47.49 & 15.86 & 43.86 \\
Wine Cabinet & 64.77 & 26.41 & 58.66 & 18.76 & 119.21 & 27.12 & 68.70 & 19.39 & 59.33 \\
TV Stand & 63.81 & 20.87 & 68.45 & 43.36 & 133.34 & 41.67 & 51.02 & 14.61 & 53.41 \\
Drawer Chest & 59.99 & 35.19 & 65.12 & 39.93 & 128.75 & 59.92 & 59.55 & 21.19 & 53.30\\
Shelf & 29.71 & 1.62 & 15.59 & 17.47 & 170.99 & 23.86 & 70.46 & 12.03 & 60.87 \\
Round End Table & 24.88 & 7.49 & 27.15 & 45.24 & 186.17 & 82.00 & 66.47 & 14.93 & 50.10 \\
Double/Queen/King Bed & 52.63 & 19.02 & 45.85 & 13.08 & 131.75 & 28.42 & 83.82 & 23.74 & 68.75 \\
Bunk Bed & 39.75 & 23.20 & 35.08 & 19.70 & 58.83 & 40.44 & 69.06 & 38.23 & 50.25 \\
Bed Frame & 50.92 & 11.20 & 41.58 & 75.69 & 359.99 & 164.11 & 75.38 & 4.62 & 36.15 \\
Single Bed & 55.32 & 16.70 & 48.72 & 11.97 & 192.22 & 24.52 & 83.86 & 16.47 & 68.04 \\
Kids Bed & 42.21 & 16.45 & 36.57 & 17.22 & 145.30 & 38.55 & 74.68 & 22.13 & 56.00 \\
Dining Chair & 40.76 & 15.87 & 40.80 & 11.73 & 109.80 & 30.77 & 86.13 & 27.18 & 69.33 \\
Lounge/Office Chair & 45.65 & 23.85 & 47.31 & 12.89 & 100.45 & 31.15 & 82.75 & 29.76 & 65.17 \\
Dressing Chair & 39.47 & 23.20 & 31.63 & 23.22 & 107.77 & 50.03 & 64.97 & 21.79 & 45.94 \\
Classic Chinese Chair & 23.77 & 13.88 & 31.90 & 21.30 & 108.05 & 52.32 &71.11 & 24.55 & 50.90 \\
Barstool & 23.32 & 6.85 & 37.43 & 20.28 & 162.84 & 57.72 & 76.95 & 14.28 & 59.33 \\
Dressing Table & 42.18 & 18.57 & 44.51 & 29.53 & 152.15 & 49.97 & 58.61 & 17.79 & 47.50\\
Dining Table & 43.07 & 10.36 & 40.39 & 56.83 & 171.71 & 85.80 & 49.81 & 16.51 & 42.71 \\
Desk & 41.41 & 12.18 & 37.92 & 67.41 & 170.40 & 96.82 & 43.51 & 16.03 & 36.12 \\
Three-Seat Sofa & 59.60 & 24.39 & 59.06 & 12.16 & 89.77 & 16.24 & 83.81 & 33.43 & 77.42 \\
Armchair & 51.27 & 33.34 & 50.63 & 16.01 & 87.83 & 33.13 & 76.77 & 32.94 & 59.18 \\
Loveseat Sofa & 56.53 & 29.01 & 57.14 & 13.31 & 72.93 & 17.77 & 81.55 & 37.11 & 75.03 \\
L-shaped Sofa & 61.79 & 20.13 & 35.21 & 9.74 & 125.71 & 29.81 & 85.34 & 28.06 & 68.38\\
Lazy Sofa & 45.21 & 33.80 & 54.93 & 17.57 & 106.72 & 30.54 & 75.89 & 30.15 & 64.28 \\
Chaise Longue Sofa & 52.57 & 21.69 & 40.85 & 19.94 & 117.47 & 34.11 & 69.22 & 26.60 & 57.92 \\
Stool & 44.74 & 39.51 & 61.92 & 20.82 & 96.18 & 39.55 & 78.61 & 39.51 & 61.92 \\
Pendant Lamp & 25.37 & 4.73 & 20.64 & 30.52 & 215.87 & 54.45 & 69.97 & 16.73 & 53.06 \\
Ceiling Lamp & 50.94 & 22.44 & 50.03 & 45.70 & 170.39 & 71.14 & 57.71& 20.14 & 46.74 \\
\hline
Mean & 47.32 & 21.32 & 46.50 & 32.35 & 144.76 & 57.33 & 65.69 & 21.42 & 53.46 \\
\hline
\end{tabular}
\caption{Numerical comparison of our several baselines for single image 3D reconstruction on our 3D-FUTURE dataset. Metrics are IoU (\%), CD ($\times 10^{-3}$, computed on 2,048 points) and F-score (thresholds is 1\%, the reconstruction volume side length defined in \citep{tatarchenko2019single}). IoU and F-score: Higher is better. CD: Lower is better.}
\label{tab:reconstruction}
\end{table*}

 \begin{figure}[th!]
\centering
\includegraphics[width=0.48\textwidth]{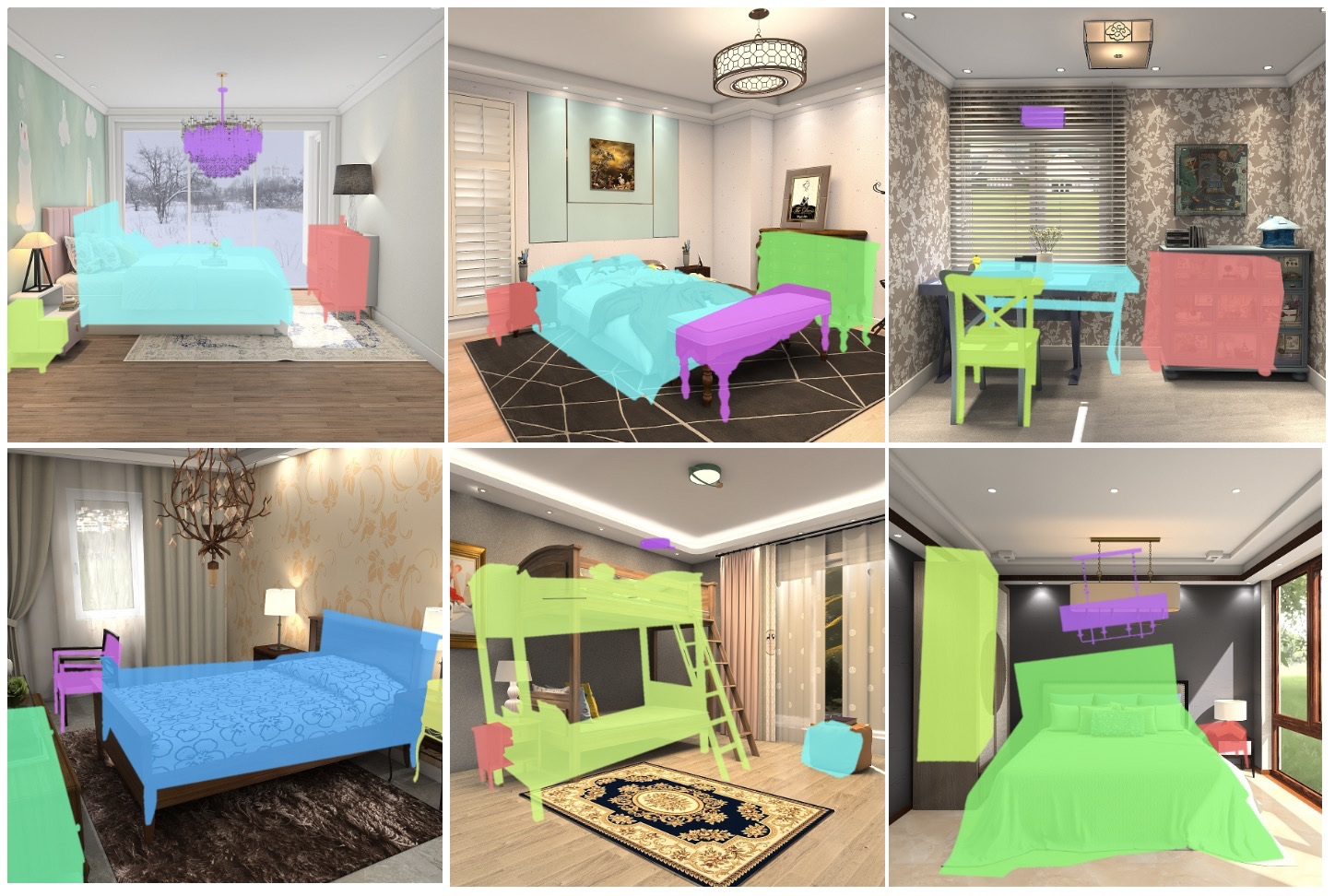}
\caption{The pose estimation results. Zoom in for better view.}
\label{fig:pose-results}
\end{figure}

In this experiment, we train the baseline using 31,444 image-shape pairs and evaluate the retrieval algorithm via the other 5,994 image-shape pairs. Then we crop the furniture instances with occlusion levels of ``NO", ``Slight" and ``Standard" from the scene images to produce the image-shape pairs. The statistics of the train and test sets are presented in Table.~\ref{tab:retrieval-sta}. We develop a  DCNN based metric learning network to study the cross-domain shape similarities, as shown in ~\ref{fig:retrieval-net}. Specifically, we first project the selected 3D shapes into 2D planes using the toolbox\footnote{https://github.com/3D-FRONT-FUTURE} to bridge the 3D and 2D gaps. Given a query image and its corresponding 3D shape, we randomly sample a negative 3D shape from the 3D pool to construct a triplet. We then feed the triples (2D images) into a ResNet-34 feature extractor and adopt a margin ranking loss to push the query image close to its corresponding 3D shape.We utilize a category classification loss and an instance classification loss \citep{wu2018unsupervised} such that the network can capture shape similarity among furniture instances. 

 \begin{figure*}[th!]
\centering
\includegraphics[width=0.98\textwidth]{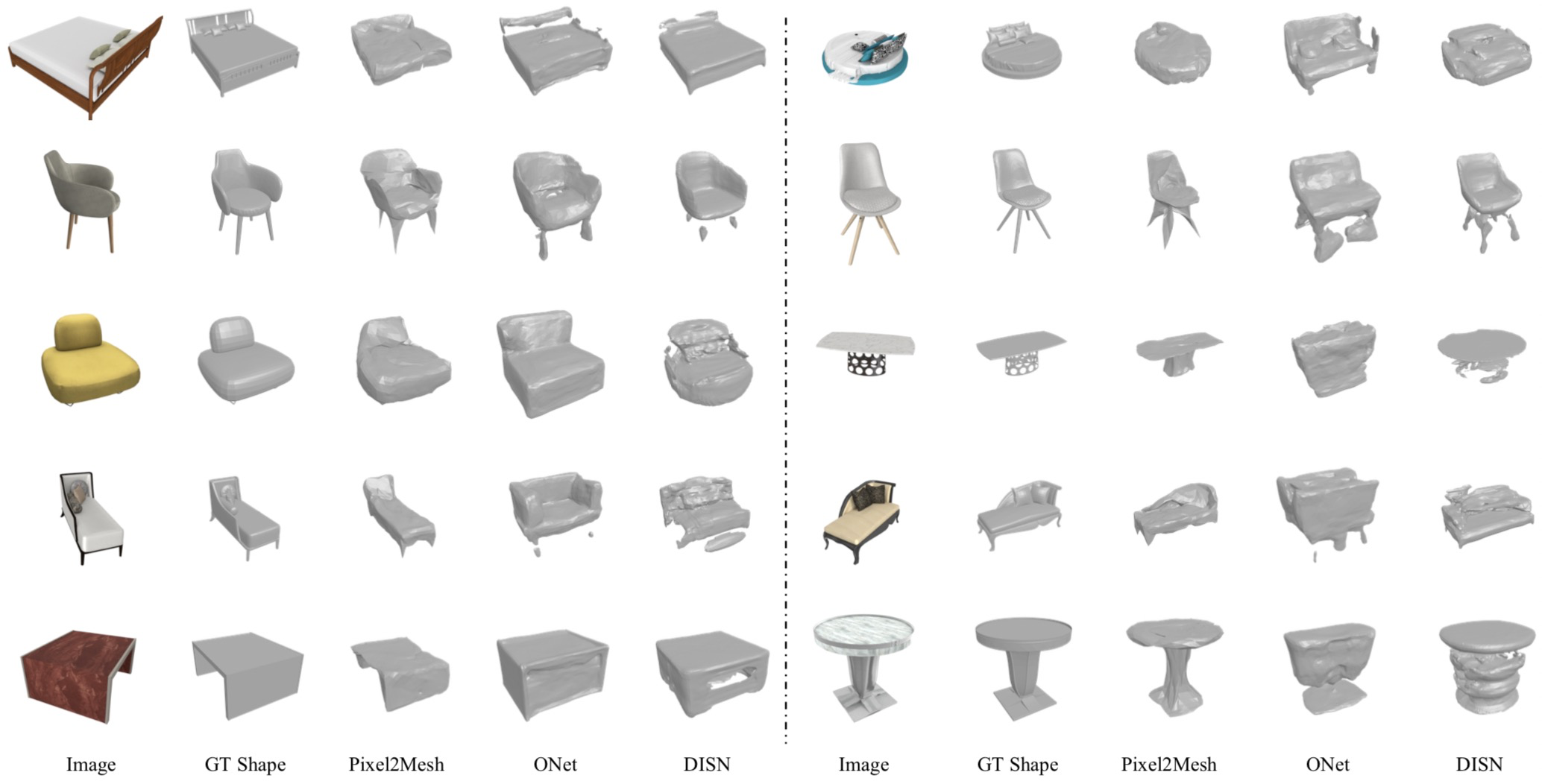}
\caption{Sample reconstruction results on our 3D-FUTURE benchmark. The SOTA methods cannot model the local geometric details.}
\label{fig:reconstruction-results}
\end{figure*}

We take TopK Recall (TopK@R) and Top5 average F-score (mean F-score) as our metrics. The latter is used to measure the retrieval sequences. The retrieval results for each category are reports in Table.~\ref{tab:retrieval}. We also show some qualitative retrieval sequences in Figure~\ref{fig:retrieval-results}. We can see that while the captured Top1@R for a large portion of categories is less than 30.0\%, the retrieval sequences seem to be visually acceptable. Besides, there is a remarkable gap between Top1@R (23.4\%) and Top3@R (40.6\%). The observations demonstrate that our large 3D pool contains many furniture with similar shape characteristics, which would provide potential opportunities for fine-grained shape retrieval studies.

\subsection{Jointly 2D Instance Segmentation and 3D Pose Estimation}

Image-based 6DoF pose estimation is a fundamental 3D vision task that can benefit many intelligent applications such as
autonomous driving, augmented reality, and robotic manipulation. Typical methods 6DoF pose estimation first build point-wise correspondences between 3D models and 2D images, followed by the Perspective-n-Point (PnP) algorithm to compute pose parameters \citep{collet2011moped,rothganger20063d}. These approaches perform well for objects with rich textures but are not robust to featureless or occluded cases. Recent works thus employ RGB-D sensors and deep learning to improve keypoints detection or directly predict 6DoF pose from images \citep{kehl2016deep,brachmann2014learning,bo2014learning,hinterstoisser2012model,xiang2017posecnn,peng2019pvnet,song2020hybridpose,tekin2018real,rad2017bb8,park2020latentfusion}. Nevertheless, the main issues such as occlusion and clutter, scalability to multiple objects, and symmetries have not been well addressed. 

Instance segmentation is the task of detecting and delineating each distinct object of interest appearing in an image. Current instance segmentation methods can be roughly categorized into two paradigms: segmentation-based methods and detection-based methods. The former category of approaches group the predicted category labels via techniques such as clustering \citep{dhanachandra2015image}, metric learning \citep{fathi2017semantic}, and watershed algorithms \citep{najman1994watershed}, to form instance segmentation results. The latter predicts the mask for region instances detected by SOTA object detectors. Methods such as Mask R-CNN series \citep{he2017mask,huang2019mask,cai2019cascade} have achieved impressive performance for daily objects. 

\begin{figure*}[th!]
\centering
\includegraphics[width=0.96\textwidth]{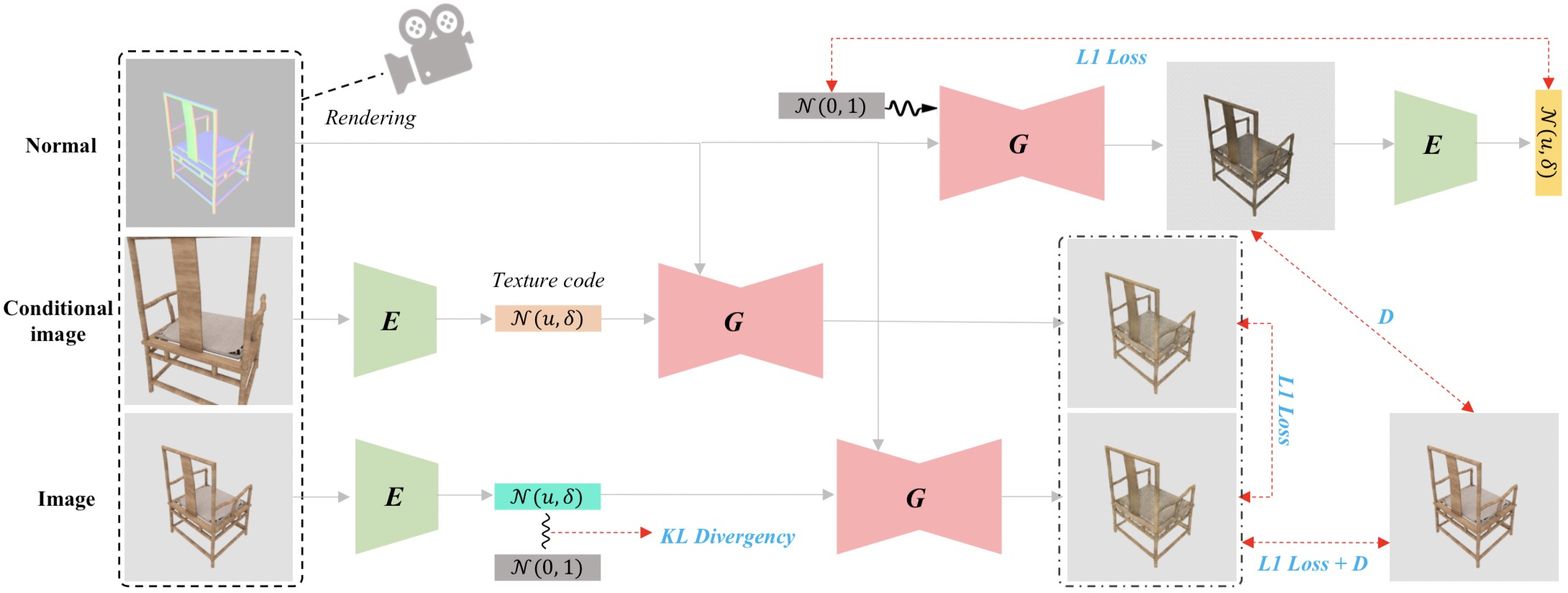}
\caption{An illustration of our BicycleGAN++ baseline. The input are rendered images from 3D shapes. \textbf{E}: Texture Encoder. \textbf{G}: Generator.}
\label{fig:bicycle-net}
\end{figure*}

In this experiment, we learn to predict instance segmentation in 2D images and estimate their 6DoF poses in a unified framework. In contrast to the well-studied benchmarks such as ObjectNet3D \citep{xiang2016objectnet3d}, PASCAL3D+ \citep{xiang2014beyond}, and Pix3D \citep{sun2018pix3d}, 3D-FUTURE encourages estimating pose parameters for multiple objects with occlusions in diverse indoor scenes. We provide 3D pose annotations for 100K+ objects in the scene images. The objects are further divided into five occlusion levels, including ``NO", ``Slight", ``Standard", ``Heavy", and ``N/A". Here, an object labeled as ``N/A" means that its corresponding 3D shape is not available, or a part of the object is out of the camera view. We train our model on the 14,761 training images and test it on the remaining 5,479 test images.

\begin{table}[th!]
\centering
\begin{tabular}{ c || c  c  | c  c  }
\hline
\multirow{2}{*}{Category} & \multicolumn{2}{c|}{Train} & \multicolumn{2}{c}{Test} \\ \cline{2-5}
& Image & Shape & Image & Shape \\
\hline
Sofa & 49,056 & 1,533 & 8,460 & 705 \\
Bed & 20,032 & 626 & 3,912 & 326 \\
Chair & 26,208 & 819 & 4,152 & 346 \\
Table & 12,320 & 385 & 2,700 & 225 \\
\hline
Total & 107,616 & 3,363 & 19,224 & 1,602 \\
\hline
\end{tabular}
\caption{The statistics of the training and test sets for the subject of texture synthesis for 3D shapes.}
\label{tab:texture-set}
\end{table}

We modify Cascade Mask-RCNN \citep{cai2019cascade,he2017mask} as our baseline. The network architecture is shown in Figure.~\ref{fig:instance-net}. Specifically, we take ResNeXt-101 \citep{xie2017aggregated} with the setting of 64-4d (group number: 64, width of group: 4) as the backbone, and adopt FPN \citep{lin2017feature} to extract the dense features. Then, we utilize a three-stage cascade architecture to perform bounding box regression and object classification. Finally, we add two branches that consist of several fully connected layers to predict the instance masks and their 6DoF poses simultaneously.  We cast rotation estimation as a viewpoint classification problem. In detail, we convert the rotation matrices to Euler angles and divide the 360-degree azimuth, 180-degree elevation, and 360-degree in-plane rotation into 18 bins, 9 bins, and 18 bins, respectively. For translation estimation, we use L1 smooth loss to regress the translation parameters directly.

For 2D instance segmentation, we report Average Precision (AP) and Average Recall (AR) over different  IoU thresholds following \citep{he2017mask}. For 3D pose estimation, we take both Average Viewpoint Precision (AVP) in PASCAL3D+ \citep{xiang2014beyond} and Average Orientation Similarity (AOS) in KITTI \citep{geiger2012we} to measure the rotation predictions as \citep{xiang2016objectnet3d}, and employ Root Mean Square Error (RMSE) to evaluate the translation predictions. In specific, we define the difference between an estimated rotation matrix $R$ and its ground truth $R_{gt}$ as $\nabla(R, R_{gt}) = \frac{1}{\sqrt{2}}\parallel \log(R^TR_{gt})\parallel_{F}$. In AVP, a correct estimation should satisfy $\nabla(R, R_{gt}) < \frac{\pi}{6}$. The cosine similarity between rotations in AOS is computed as $cos(\nabla(R, R_{gt}))$. 

We present the instance segmentation and pose estimation results in Table~\ref{tab:pose-estimation}. Here, the metrics for camera poses are with respect to AP and AR, where the IoU thresholds range from 0.5 to 0.95. For instance segmentation, our baseline captures a mean AP of 0.55 on 3D-FUTURE. The score is at a similar level to those reported on the MSCOCO leaderboard achieved by recent SOTA methods. For 3D pose estimation, our baseline yields a mean AVP of 43\%. Besides, as analyzed in \citep{xiang2016objectnet3d}, AP is an upper bound of AOS. This means the closer AOS is to AP, the more accurate the rotation estimation is. By showing the gaps between AOS and AP in Figure~\ref{fig:histogram-aos}, we can see that the estimated rotation (0.43) can be further improved. From the observations, we conclude that most objects' 3D poses are not well modeled in our challenging setting. This suggests that researchers may need to carefully study 3D pose estimation with different levels of occlusions based on 3D-FUTURE. Some qualitative results are shown in Figure~\ref{fig:instance-seg} and Figure~\ref{fig:pose-results} to further justify our conclusions.

\setlength\tabcolsep{5.0pt}
\begin{table*}[th!]
\centering
\begin{tabular}{ c || c  c  c  c | c  c  c  c }
\hline
\multirow{2}{*}{Category} & \multicolumn{4}{c|}{Texture Field} & \multicolumn{4}{c}{BicyleGAN++} \\ \cline{2-9}
& FID & SSIM & L1 & Feat1 & FID & SSIM & L1 & Feat1 \\
\hline
Sofa & 22.01 & 0.959 & 0.013 & 0.168 & 10.01 & 0.951 & 0.019 & 0.146 \\
Bed & 37.22 & 0.924 & 0.024 & 0.190 & 18.06 & 0.916 & 0.030 & 0.172 \\
Chair & 15.36 & 0.951 & 0.017 & 0.131 & 10.65 & 0.941 & 0.022 & 0.120 \\
Table & 29.45 & 0.964 & 0.011 & 0.149 & 21.78 & 0.958 & 0.016 & 0.137 \\
\hline
mean & 26.01 & \textbf{0.952} & \textbf{0.016} & 0.160 & \textbf{15.12} & 0.942 & 0.022 & \textbf{0.144} \\
\hline
\end{tabular}
\caption{Quantitative Evaluation using the FID, SSIM, \emph{L1}, and \emph{Feat1} metrics. FID, \emph{L1}, \emph{Feat1}: lower is better. SSIM: higher is better.}
\label{tab:texture}
\end{table*}

\begin{figure*}[t]
\centering
\includegraphics[width=0.98\textwidth]{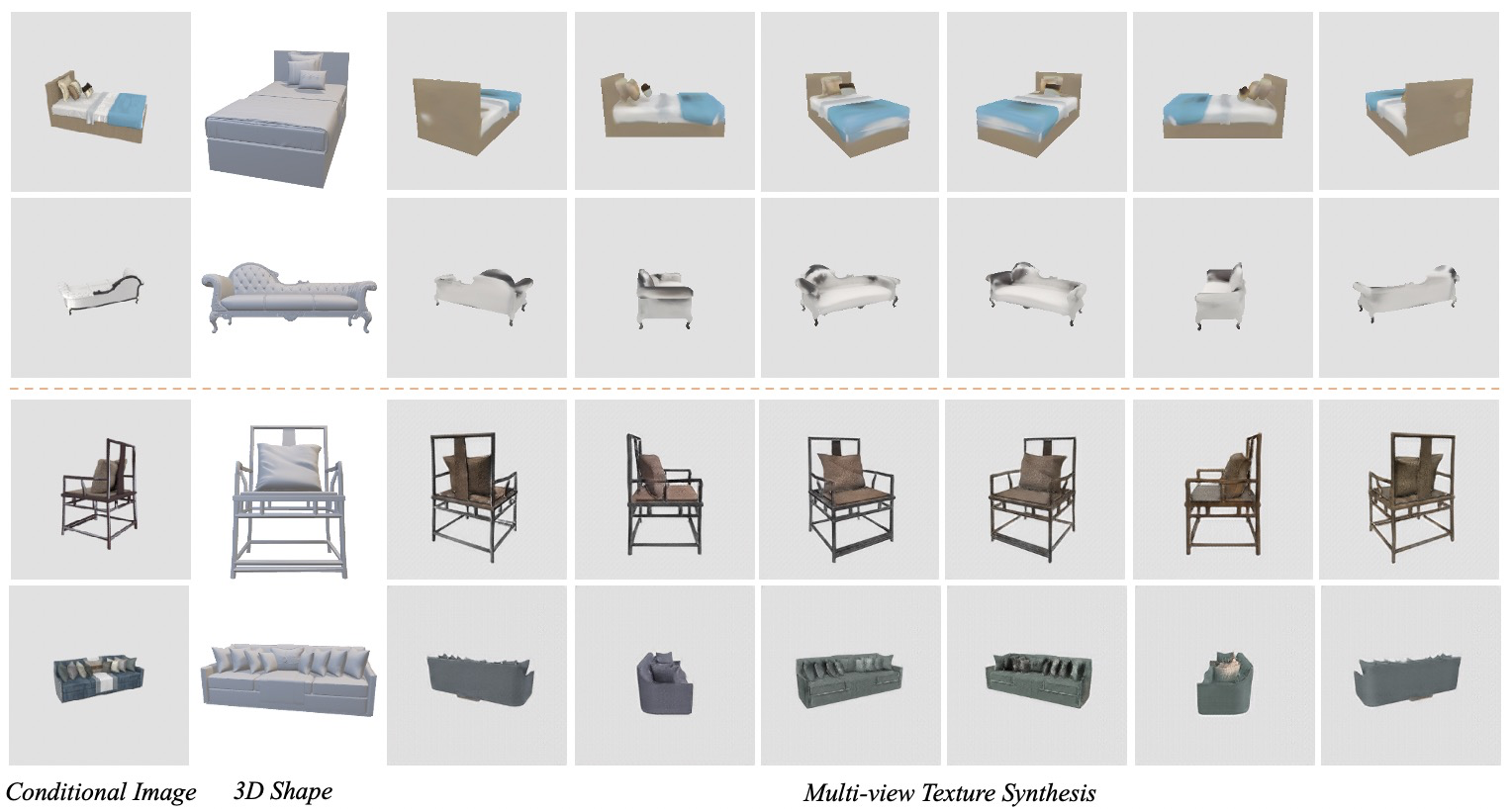}
\caption{The multi-view texture synthesis results. Top: Texture Fields \citep{oechsle2019texture}. Bottom: Our BicycleGAN++ based on BicycleGAN \citep{zhu2017toward}.}
\label{fig:texture-multiview}
\end{figure*}


\subsection{Single-View 3D Object Reconstruction}
Inferring 3D structure from a single image has been an active research area for a long time. In the supervised setting, traditional methods investigated shape from shading \citep{durou2008numerical,zhang1999shape} and defoce \citep{favaro2005geometric} to reason the visible parts of objects. Leveraging on large-scale shape repositories, various works examined deep architectures to produce shapes in 3D volume \citep{choy20163d}, point cloud \citep{fan2017point}, and mesh surface \citep{groueix2018} directly. Recently, several SOTA methods recovered 3D meshes from initializations using shape deformation based on deep networks \citep{wang2018pixel2mesh}. In the unsupervised setting, 3D recovery has been recast as a 2D image reconstruction progress of unobserved views with differentiable rendering \citep{liu2019soft,chen2019learning}.

In this paper, we examine several SOTA reconstruction algorithms as the baselines, including ONet \citep{mescheder2019occupancy}, Pixel2Mesh \citep{wang2018pixel2mesh}, and DISN \citep{xu2019disn}. We report the widely studied Intersection over Union (IoU), Chamfer Distance (CD), and F-score to evaluate these approaches on 3D-FUTURE. We refer \citep{xu2019disn} for the definitions of these metrics. We randomly render 24 different view images each model for training and a random view image for testing. The resolution of each image is $256\times 256$.  As shown in Table~\ref{tab:reconstruction} and {Figure~\ref{fig:reconstruction-results}}, Pixel2Mesh is more robust in general 3D object reconstruction.  However, all the SOTA methods cannot recover good-quality shapes when the 3D shapes contain many geometric details.

\begin{figure*}[t]
\centering
\includegraphics[width=0.98\textwidth]{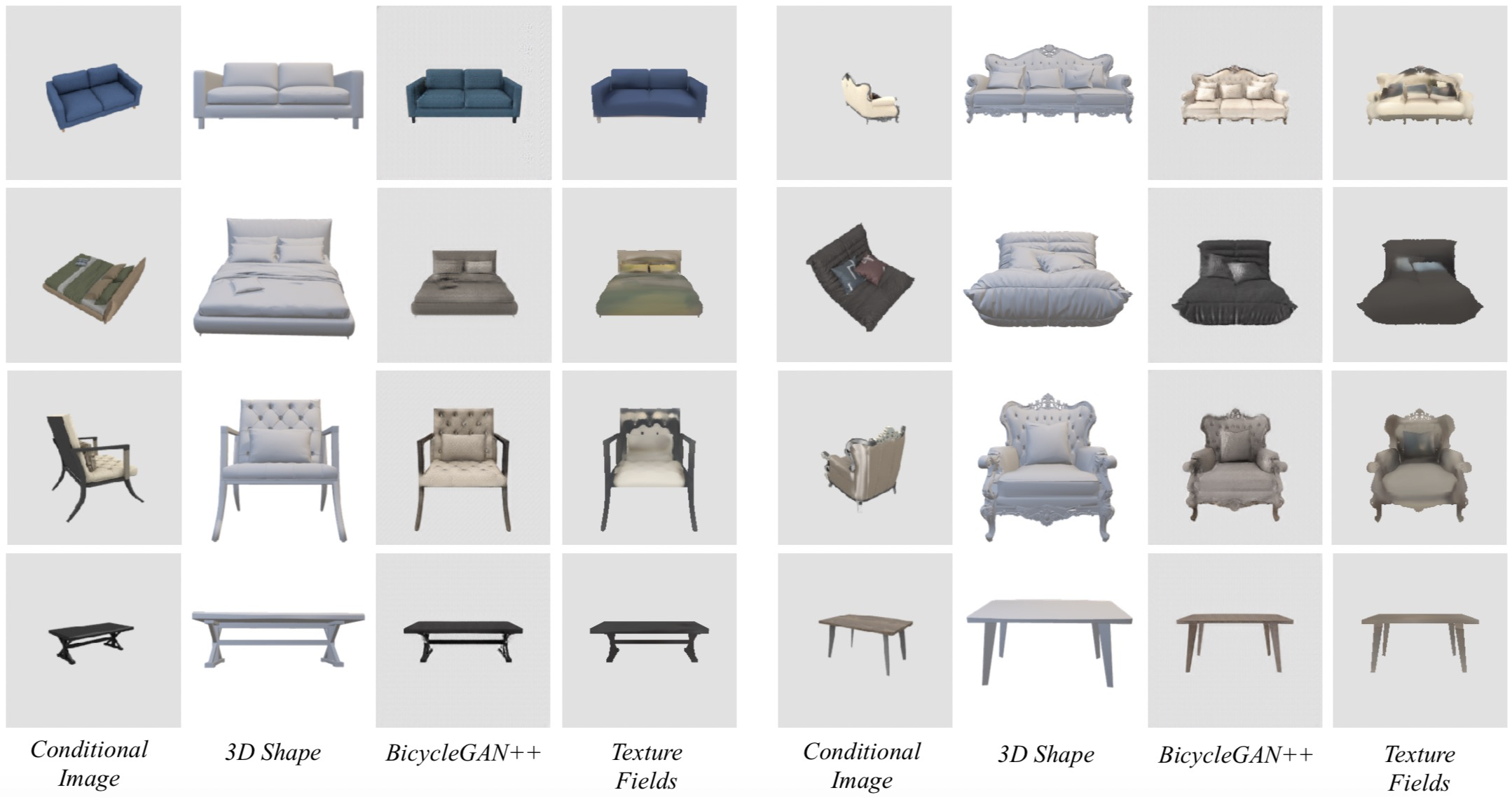}
\caption{A quantitative comparison between Texture Fields and BicycleGAN++ for conditional texture synthesis.}
\label{fig:texture-results}
\end{figure*}


\subsection{Texture Synthesis For 3D Shapes}

Unlike geometry reconstruction, texture reconstruction of 3D objects has received less attention from the community. Previous works studied the subject by learning colored 3D reconstruction on voxels or point clouds \citep{sun2018im2avatar,tulsiani2017multi} based on view synthesis and multi-view geometry. While voxel representations are limited to the low resolutions, point representations are sparse and thus ignore geometric details. Recent approaches alternatively learned a 2D texture atlas (UV mapping) for 3D meshes to map a point on the shape manifold to a pixel in the texture atlas. These methods mainly take advantage of differentiable rendering to recast the problem as an unobserved view synthesis problem \citep{raj2019learning,oechsle2019texture}.

Existing 3D repositories contain less dreamlike or uninformative textures and cannot support high-quality texture recovery studies. In contrast, 3D-FUTURE provides furniture shapes with informative textures, which are widely used in industrial productions. We examine two baselines for texture synthesis, \emph{i.e.,} Texture Fileds \citep{oechsle2019texture} and a novel BicycleGAN++ method. Here, BicycleGAN++ extends BicycleGAN \citep{zhu2017toward} for texture synthesis. An illustration of the network is shown in Figure~\ref{fig:bicycle-net}. In specific, we incorporate a texture encoder such that the learned model can perform controllable texture synthesis. Importantly, by enlarging the weights of the reconstruction losses and introducing a texture consistency loss, we find that the produced multi-view textured images will show preferable consistency in overlap regions.  

We conduct experiments on four super-categories, including Sofa, Bed, Chair, and Table. The details of our train and test splits are reported in Table~\ref{tab:texture-set}. We randomly render 32 views of images for each shape to enlarge the training set. For each baseline, we first train them on the whole train set and then perform category-specific fine-tuning. Following \citep{oechsle2019texture}, we use structure similarity image metric (SSIM) \citep{wang2004image}, \emph{L1}, Frechet Inception Distance (FID) \citep{heusel2017gans}, and \emph{Feat1} as our metrics to evaluate the quality of the synthetic texture. Here, \emph{L1} is the L1 distance between the ground-truth view rendering and the produced textured image under the same viewpoint. \emph{Feat1} is a global perceptual measure operated on the Inception-net \citep{szegedy2015going} feature space using the L1 distance.
As shown in Table~\ref{tab:texture}, while BicycleGAN++ earns higher scores on FID and \emph{Feat1}, Texture Fields performs better in terms of SSIM and L1, indicating that BicycleGAN++ produces more realistic images with higher quality and Texture Fields focuses more on structured texture details. We also give some qualitative results in Figure~\ref{fig:texture-multiview} and Figure~\ref{fig:texture-results}. We can see that BicylcGAN++ can only learn the main color information while largely ignores the semantic parts of objects. Texture Fields can partially preserve the structured texture details but produces dreamlike textures. These observations demonstrate that achieving visually appealing texture recovery for 3D meshes is still very challenging,  especially for the industrial 3D shapes with informative texture details.

\section{Conclusion}
\label{sec:conclusion}
In this paper, we have built the large-scale 3D-FUTURE benchmark specific to the household scenario with rich 3D and 2D annotations. 3D-FUTURE contains 20,240 realistic synthetic images and 9,992 high-quality 3D CAD furniture shapes. The exciting features include but are not limited to the exhausting interior designs by experienced designers, photo-realistic renderings, 2D-3D alignments, and most significantly the industrial 3D furniture shapes with informative textures. We conduct several experiments to show the remarkable properties of 3D-FUTURE. The experiments can serve as baselines for future research using our database. We hope that 3D-FUTURE can facilitate innovative research on high-quality 3D shape understanding and generation, bring new research opportunities for 3D vision, and build a bridge between academic study and 3D industrial applications.




%
%


\bibliographystyle{spbasic}
\bibliography{egbib}
\end{document}